%% file: main.tex
\documentclass[accepted]{uai2025}

\usepackage[american]{babel}

\usepackage{acronym}
\usepackage{verbatim}
\usepackage{pifont}

\usepackage{natbib} %
\input{math_commands}

\newcommand{\perm}[1][]{\tau_{#1}}

\newcommand{\cmark}{\ding{51}}%
\newcommand{\xmark}{\ding{55}}%

\acrodef{EI}[EI]{expected improvement}
\acrodef{PI}[PI]{probability of improvement}
\acrodef{KG}[KG]{knowledge gradient}
\acrodef{GD}[GD]{gradient descent}
\acrodef{RS}[RS]{random search}
\acrodef{DM}[DM]{deterministic selection}
\acrodef{AF}[AF]{acquisition function}
\acrodef{MES}[MES]{max-value entropy search}
\acrodef{ES}[ES]{entropy search}
\acrodef{PES}[PES]{predictive entropy search}
\acrodef{JES}[JES]{joint entropy search}
\acrodef{BO}[BO]{Bayesian optimization}
\acrodef{HDBO}[HDBO]{high-dimensional Bayesian optimization}
\acrodef{UCB}[UCB]{upper confidence bound}
\acrodef{OTSD}[OTSD]{observation traveling salesman distance}
\acrodef{OE}[OE]{observation entropy}
\acrodef{DoE}[DoE]{design of experiments}
\acrodef{EETO}[EETO]{exploration-exploitation trade-off}
\acrodef{GP}[GP]{Gaussian process}
\acrodef{TCD}[TCD]{total center distance}
\acrodef{MAB}[MAB]{multi-armed bandit}
\acrodef{MDP}[MDP]{Markov decision process}
\acrodefplural{MDP}[MDPs]{Markov decision processes}
\acrodef{RL}[RL]{reinforcement learning}
\acrodef{EA}[EA]{evolutionary algorithm}
\acrodef{TSP}[TSP]{traveling salesman problem}
\acrodef{PDF}[PDF]{probability density function}
\acrodef{KL}[KL]{Kozachenko-Leonenko}
\acrodef{TR}[TR]{trust region}
\acrodef{TS}[TS]{Thompson sampling}
\acrodef{RAASP}[RAASP]{random axis-aligned subspace perturbations}

\title{Exploring Exploration in Bayesian Optimization}

\author[1]{\href{mailto:<leonard.papenmeier@cs.lth.se>?Subject=Your paper}{Leonard Papenmeier*}{}}
\author[2]{Nuojin Cheng*}
\author[2]{Stephen Becker}
\author[1, 3]{Luigi Nardi}
\affil[1]{%
    Department of Computer Science\\
    Lund University\\
    Lund, Sweden
}

\affil[2]{%
    Department of Applied Mathematics\\
    University of Colorado Boulder\\
    Boulder, Colorado, USA
  }

\affil[3]{%
    DBtune
}  
\begin{document}
\maketitle
\begin{abstract}
    A well-balanced exploration-exploitation trade-off is crucial for successful \aclp{AF} in \acl{BO}.
    However, there is a lack of quantitative measures for exploration, making it difficult to analyze and compare different \aclp{AF}.
    This work introduces two novel approaches -- \acl{OTSD} and \acl{OE} -- to quantify the exploration characteristics of \aclp{AF} based on their selected observations.
    Using these measures, we examine the explorative nature of several well-known \aclp{AF} across a diverse set of black-box problems, uncover links between exploration and empirical performance, and reveal new relationships among existing \aclp{AF}.
    Beyond enabling a deeper understanding of acquisition functions, these measures also provide a foundation for guiding their design in a more principled and systematic manner.
\end{abstract}

\section{Introduction}
\label{sec:intro}
\def\thefootnote{*}\footnotetext{Equal contribution.}\def\thefootnote{\arabic{footnote}}
\ac{BO} is a widely used method to maximize black-box functions. 
Given a function $f:\mathcal{X}\to\mathbb{R}$, \ac{BO} guides the optimization process by sequentially constructing probabilistic surrogates -- typically a \ac{GP} -- and selecting new sampling points by maximizing an \ac{AF} $\alpha:\mathcal{X}\to\mathbb{R}$.
The \Ac{AF} is a key component for any \ac{BO} algorithm that chooses the point to evaluate next.
As \ac{BO} is typically used for problems where the underlying function is expected to be multi-modal, it is crucial that the \ac{AF} exhibits \emph{explorative} behavior, allowing it to discover different modes of the objective function, but also \emph{exploitative} behavior, allowing it to focus on finding the optimum in a promising region of the search space $\mathcal{X}$.
In short, a successful \ac{AF} should exhibit a good \acf{EETO} that balances these desiderata.

\begin{figure}[tb]
	\centering
	\includegraphics[width = .85\linewidth]{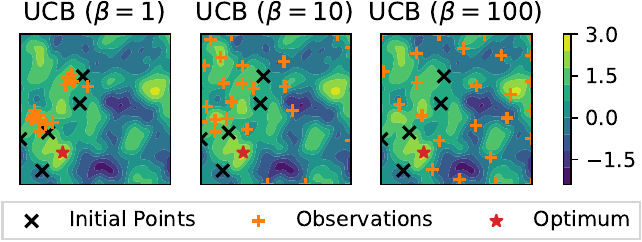}
	\caption{
    Observations of \acs{UCB} with various $\beta$ values ($1$, $10$, and $100$) on a two-dimensional \ac{GP}-prior sample reveal the explorative behavior for different $\beta$. The black crosses are initial points, the orange plus signs are the observations of the \ac{BO} phase, and the red star is the optimal location.
    }
	\label{fig:intro_ucb}
\end{figure}

It is widely recognized in the \ac{BO} community that different \acp{AF} exhibit varying degrees of explorative preference, and some \acp{AF} include parameters that allow explicit control over this behavior. For example, \ac{UCB}~\citep{srinivas2010gaussian} employs a parameter $\beta$ to regulate the level of exploration. Fig.~\ref{fig:intro_ucb} illustrates an example of \ac{UCB} with various $\beta$ values applied to a two-dimensional GP prior with a length scale of $0.1$. In this experiment, five initial points were fixed, and observations were collected over 25 iterations for each $\beta$ value. A larger $\beta$ produces a more dispersed layout of observations, indicating increased exploration. Similar behavior is observed in other acquisition functions, such as weighted expected improvement~\citep{sobester2005design} and $\epsilon$-greedy strategies~\citep{sutton2018reinforcement}. However, when comparing \acp{AF} from different families -- such as \ac{UCB} versus weighted expected improvement or $\epsilon$-greedy strategies -- assessing their exploration preferences becomes challenging, as no universally accepted metric exists to quantify these characteristics. Understanding the exploration tendencies of different \acp{AF} is important, as this knowledge influences the selection of appropriate \acp{AF} for real-world problems, especially when a specific level of exploration is required.

In this work, we fill this gap by proposing two novel quantities to quantify the level of exploration.
We make the following contributions:
\begin{itemize}[wide, leftmargin=*]
    \item We propose two novel means for quantifying exploration named \ac{OTSD} and \ac{OE}. 
    The first quantity is based on the total Euclidean distance of a traveling salesman tour connecting all the observation points in the search space, while the second adopts an information-theoretic approach and uses the empirical differential entropy of the observations.
    \item We introduce the first empirical taxonomy of acquisition function exploration based on these new methods, demonstrating their effectiveness in capturing exploration behavior.
    \item 
    We provide an extensive evaluation across a diverse set of low- and high-dimensional synthetic and real-world benchmarks, demonstrating the exploration behavior of popular acquisition functions and their extensions. \ac{OTSD} and \ac{OE} strongly correlate in benchmark problems, cross-validating their reliability.
\end{itemize}

\section{Background and Related Work}
\label{sec:related-work}
\subsection{Gaussian Processes}
A \ac{GP} is a stochastic process that models an unknown function. 
It is characterized by the property that any finite set of function evaluations follows a multivariate Gaussian distribution. 
Assuming that the prior has a zero mean, a \ac{GP} is uniquely determined by the current observations $\mathcal{D}_t\coloneqq\{(\bm x_i, y_{\bm x_i})\}_{i=1}^t$ and the kernel function $\kappa(\bm x, \bm x')$. 
Given these, at stage $t$, the predicted mean of $y_{\bm{x}}$ at a new point $\bm x$ is $\mu_t(\bm x) = \bm\kappa_t(\bm x)^T (\bm K_t)^{-1} \bm y_t$, and the predicted covariance between points $\bm x$ and $\bm x'$ is $\textup{Cov}_t(\bm x, \bm x') = \kappa(\bm x, \bm x') - \bm\kappa_t(\bm x)^T (\bm K_t+\tilde{\sigma}\bm I)^{-1} \bm\kappa_t(\bm x')$,
where $[\bm\kappa_t(\bm x)]_i = \kappa(\bm x_i, \bm x)$, $[\bm y_t]_i = y_{\bm x_i}$, $\tilde{\sigma}$ is noise level, and $[\bm K_t]_{i, j} = \kappa(\bm x_i, \bm x_j)$. At point $\bm x$, the \ac{GP} posterior $y_{\bm x}\sim\mathcal{N}(\mu_t(\bm x), \sigma_t(\bm x))$, where $\sigma_t(\bm x)\coloneqq \textup{Cov}_t(\bm x, \bm x)$; see \citet{rasmussen2006gaussian} for more details.

\subsection{Acquisition Functions}
\paragraph{One-step.} 
One commonly used \ac{AF} for \ac{BO} is \acf{EI}~\citep{jones1998efficient}, defined as $ \alpha_{\text{EI}}(\bm{x}) \coloneqq \mathbb{E}\bigl[\max\{y_{\bm{x}} - y_t^\ast, 0\}\bigr]$, where $y_t^\ast$ denotes the current-best (i.e., incumbent) observation. A closely related function is \ac{PI}~\citep{jones2001taxonomy}, which considers only the probability that a new observation $y_{\bm{x}}$ exceeds the incumbent $y_t^\ast$, without accounting for the magnitude of improvement: $ \alpha_{\text{PI}}(\bm{x}) \coloneqq \mathbb{P}\bigl[y_{\bm{x}} > y_t^\ast\bigr]$. Both \ac{EI} and \ac{PI} have closed-form expressions and are therefore computationally efficient. Similarly, \ac{UCB} is defined as $ \alpha_{\text{UCB}}(\bm{x}) \coloneqq \mu_t(\bm{x}) + \sqrt{\beta_t}\,\sigma_t(\bm{x})$, where $\beta_t$ is a parameter that balances exploration and exploitation. In contrast, information-theoretic \acp{AF} select the next sampling point to reduce uncertainty about a particular property of the optimum -- whether its location as in \ac{ES} or \ac{PES}~\citep{hennig2012entropy, hernandez2014predictive}, its value as in \ac{MES}~\citep{wang2017max}, or both as in \ac{JES}~\citep{hvarfner2022joint, tu2022joint}. 
\ac{TS} implicitly balances the exploration-exploitation trade-off by maximizing posterior samples from a Gaussian process whose accuracy improves as more observations are incorporated \citep{bijl2016sequential}. However, \ac{TS} has been criticized for being overly explorative.
To address this, several approaches have been proposed to make it less explorative, including \acp{TR} that restrict the space over which the \ac{AF} is maximized to a subregion of $\mathcal{X}$~\citep{eriksson2019scalable} and \ac{RAASP} sampling that only considers points close to the incumbent observation for the maximization of the \ac{AF}~\citep{rashidi2024cylindrical}.

\paragraph{Multi-step.}
One common property of all aforementioned \acp{AF} is that they assume that the next evaluation will be the last, i.e., they greedily maximize the simple or inference regret for the next iteration, assuming no more evaluations will be performed~\citep{wang2017max}.
In contrast, multi-step \acp{AF}~\citep{ginsbourger2010towards,wu2019practical,jiang2020efficient} consider the impact of the current choice for future evaluations: $\alpha_{\text{MS}}(\bm{x}) = v_1(\bm{x}|\mathcal{D}_t)+\mathbb{E}_y\left[\max_{\bm{x}_2}\left(v_1(\bm{x}_2\vert\mathcal{D}_t\cup\left\{(\bm{x},y_{\bm x})\right\}+\mathbb{E}_{y_2}\left[\ldots\right]\right)\right]$,
where $v_1(\bm{x})$ is the one-step marginal value of $\bm{x}$, e.g., the expected improvement upon observing $\bm{x}$.
See~\cite{jiang2020efficient} for details.
Multi-step \acp{AF}, such as \ac{KG}~\citep{frazier2009knowledge}, are computationally expensive since the expectations of $\alpha_{\text{MS}}$ must be approximated with Monte-Carlo methods and, therefore, are often limited to one lookahead step even though the theoretical framework can usually be extended to arbitrarily many lookahead steps~\citep{jiang2020efficient}.
At the same time, they are fundamentally different from previous \acp{AF} and may be characterized by a unique \ac{EETO}~\citep{wu2019practical}.

\paragraph{Batch.}

Batching is a technique used where multiple function evaluations can be performed in parallel.
Instead of re-conditioning the \ac{GP} after a single new observation, one observes $q$ points in parallel.
Batching requires modifications of the \ac{AF} to ensure that a batch contains a diverse set of candidates.
One strategy for batch \ac{BO} is using multi-point \acp{AF} that estimate the improvement of some utility upon observing $q$ new points~\citep{wang2020parallel}.
Other approaches include \emph{local penalization}~\citep{gonzalez2016batch} that repels points from regions around points already included in the batch.

\subsection{Expressions of Exploration}

\paragraph{Quantifying Exploration.}
Although achieving a good balance between exploration and exploitation is widely recognized as a crucial component of an effective acquisition function, the \ac{BO} community still lacks a simple and convenient metric for quantifying exploration. Several related metrics have been proposed -- such as \ac{TCD}~\citep[Fig.~6]{eriksson2019scalable} and incumbent distance~\citep[Fig.~20]{hvarfner2024vanilla} -- but each has notable shortcomings. For instance, \ac{TCD} focuses solely on the movement of the incumbent, and it fails when the first sample lands on the optimum (yielding a \ac{TCD} of zero) or when the search oscillates between multiple optima. Similarly, incumbent distance measures the distance between the next query and the incumbent but does not reliably capture exploration, as repeatedly querying the opposite corner of the space can yield high values without representing meaningful explorative behavior.

Another approach employs Pareto analysis- 
to interpret the \ac{EETO}~\citep{bischl2014moi,feng2015multiobjective,vzilinskas2019bi,de2021greed}. In this framework, exploration and exploitation are treated as distinct objectives, with the utility function depending on both the predicted mean $\mu$ and variance $\sigma$. Under this interpretation, many \acp{AF}, such as \ac{PI}, \ac{EI}, and \ac{UCB}, can be accommodated. However, more sophisticated \acp{AF}, particularly those based on information-theoretic principles, do not conform to this framework.

\paragraph{Quantifying Exploration in Evolutionary Algorithms.}
The measurement of the exploration extends beyond Bayesian optimization. In evolutionary algorithms (EA)~\citep{eiben2015introduction}, this quantity is evaluated from two perspectives: genotypic and phenotypic~\citep{vcrepinvsek2013exploration}. Genotypic measures assess diversity in the input space using tree-based techniques~\citep{burke2002advanced,vcrepinvsek2011analysis}, Euclidean distance quantities~\citep{mcginley2011maintaining}, entropy-based methods~\citep{misevivcius2011generation}, and individual-population similarity indices~\citep{inoue2015analyzing}. In contrast, phenotypic measures evaluate diversity in terms of fitness or performance, employing distance-based methods~\citep{chaiyaratana2007effects}, entropy-based measures~\citep{adra2010diversity, turkey2014model}, and local-attraction metrics~\citep{jerebic2021novel}. Notably, the distance-based approaches mentioned above typically measure the distance from each individual to the population's center rather than the aggregate distance connecting all individuals, distinguishing them from our method. Additionally, entropy-based approaches have mainly focused on low-dimensional discrete domains -- often using binning methods to approximate entropy -- whereas our proposed techniques target high-dimensional continuous domains.

\paragraph{Quantifying Exploration in Reinforcement Learning.}
In \ac{RL}, whether in simpler settings like \acp{MAB}, a special case of \acp{MDP} where the state remains constant, or in more complex \ac{MDP} scenarios, an agent must balance exploration and exploitation to achieve long-term benefits.

In \ac{MAB} problems, one measure of exploration is tracking the frequency with which each action (arm) is selected ~\citep{kuleshov2014algorithms}. For general \ac{RL} tasks, both states and actions must be considered. 
Some methods promote exploration by maximizing the entropy of the action distribution~\citep{williams1991function,ahmed2019understanding} to encourage the agent to try diverse actions. 
Curiosity-driven exploration, often quantified using information gain~\citep{sun2011planning,houthooft2016vime}, provides another approach to exploration in the state-action space. 
Existing entropy-based exploration methods in \ac{RL} often rely on parametric methods, like variational inference, which leverage prior knowledge of the underlying density functions (e.g., for actions) but may be inaccurate when the parametric assumptions fail. In \ac{BO}, we do not have access to the true density of observations, so we adopt a non-parametric approach for density estimation. Consequently, our method is not directly comparable to previous entropy-based approaches in \ac{RL} that depend on specific parametric families.

\section{Quantifying Exploration}
\label{sec:metrics}
Motivated by the \cite{dictionary} dictionary definition of \emph{exploration} as ``the activity of travelling to and around a place, especially one where you have never been [\ldots] before, in order to find out more about it'', we define exploration in the context of black-box optimization. 
\begin{definition}[Exploration]
    The activity of sampling in a region of the search space, especially one that has never been sampled before, to learn more about a global optimum.
\end{definition}
Quantifying the exploration preferences of \acp{AF} is crucial for understanding their behavior and for developing an effective \ac{AF} portfolio to achieve better performance. In this section, we first summarize existing knowledge on the exploration tendencies of different acquisition functions and then propose two key methods to quantify exploration.

\subsection{Analysis of Tribal Knowledge}
\label{sec:analysis_of_tribal_knowledge}
The \ac{EI} and \ac{PI} \acp{AF} have been shown to explore relatively little~\citep{de2021greed}, with \ac{PI} being even less explorative than \ac{EI}~\citep{benjamins2022pi}. In contrast, the \ac{KG} \ac{AF} -- which can be seen as a generalization of \ac{EI}~\citep[p.\ 12]{wu2017knowledge} -- has been reported to be more explorative than \ac{EI}~\citep[p.\ 89]{frazier2009knowledge}. Information-theoretic acquisition functions are generally considered on the explorative side~\citep{hernandez2014predictive}, although they can be surpassed by \ac{UCB} with a high $\beta$ value. At the extremes of the spectrum, \ac{RS} samples points uniformly at random throughout the domain $\mathcal{X}$, while \ac{DM} always selects the same fixed point; both completely disregard the probabilistic surrogate model. Empirical findings on the explorative behavior of \acp{AF} can be broadly summarized by the following informal ordering: $ \text{RS} \succeq \text{UCB (high $\beta$)} \succeq \text{Information Theoretic}\,\raisebox{.5pt}{\textcircled{\raisebox{-.9pt}{\small ?}}}\,\text{KG} \succeq \text{EI} \succeq \text{PI} \succeq \text{UCB (low $\beta$)} \succ \text{DM} $. Finally, although \ac{TS} is known to be explorative~\citep{do2024epsilon}, its exact placement in this ranking remains unclear. This ordering reflects a general quantitative understanding within the \ac{BO} community; however, the relative explorative behavior of acquisition functions may vary depending on the specific problem setting. In particular, the relationship between information-theoretic acquisition functions and \ac{KG} remains uncertain, as indicated by the question mark.

\subsection{Exploration Methods}

We introduce two quantities to evaluate the exploration behavior of different black-box optimization methods based on the locations of their observations in the search space.
\paragraph{Observation Traveling Salesman Distance (OTSD).}
\Ac{OTSD} quantifies the minimum Euclidean distance required to connect all observation points by formulating the problem as a \ac{TSP}. 
Given a set of $t$ observation points $X_t\coloneqq\{\bm{x}_i\}_{i=1}^t$ from $\mathcal{D}_t$, \ac{OTSD} is defined as the total length of the shortest possible route that visits each observation point exactly once and returns to the starting point. 
Mathematically, it is expressed as:
\begin{equation}
\begin{aligned}
\text{OTSD}(X_t)\coloneqq \min_{\perm \in S_t} \left( \sum_{i=1}^{t} \lVert \bm{x}_{\perm(i)} - \bm{x}_{\perm(i+1)} \rVert \right),
\end{aligned}
\end{equation}
where $\perm$ is a permutation of $\{1, 2, \dots, t\}$ representing a tour that visits all points with $\perm(t+1)\coloneqq\perm(1)$, $\lVert \cdot \rVert$ denotes the Euclidean distance, and $S_t$ is the set of all possible permutations of $t$ elements. 
\ac{OTSD} increases monotonically with the number of observations.

Since the \ac{TSP} is NP-hard, we approximate the solution using an insertion heuristic method~\citep{rosenkrantz1974approximate}, which has a time complexity of $\mathcal{O}(dT^2)$ for $T$ observations in $d$ dimensions, and the worst-case tour length is guaranteed to be at most twice the optimal distance. This approach conveniently allows us to track the \ac{OTSD} for each $t \leq T$. %
The pseudocode for calculating \ac{OTSD} is in Algorithm~\ref{alg:otsd}.

\begin{algorithm}[ht]
    \DontPrintSemicolon
    \caption{OTSD Insertion Heuristic\label{alg:otsd}}
    \KwIn{Observation locations $X_t = \{\bm{x}_1, \bm{x}_2, \dots, \bm{x}_t\}$}
    \KwOut{Estimated TSP distance, OTSD($X_t$)}
    
    \begin{algorithmic}[1]
        \STATE Compute pairwise distances: $D(i,j) = \|\bm{x}_i - \bm{x}_j\|$ for all $i,j \in \{1, \dots, t\}, i\neq j$.
        \STATE Initialize the permutation $\perm: \{1\} \to \{1\}$ that represents the (trivial) tour order on one point, $\bm{x}_1$.  \\ %
        For a tour on $k$ points, define $\perm[i] \coloneqq\perm(i \text{ mod } k)$.
        \STATE Initialize $\text{OTSD} \gets 0$
        
        \FOR{$k = 2$ to $t$}
            \STATE For each consecutive pair in the tour $\perm$ on points $\{\bm{x}_1,  \dots, \bm{x}_{k-1}\}$, compute the insertion cost for placing $\bm{x}_k$ between them:  $\forall i=1,\ldots,k-1$, \\
            $\Delta C(i) \coloneqq D(\perm[i],k) + D(k, \perm[i+1]) - D(\perm[i], \perm[i+1])$            
            \STATE Identify the insertion point $i^\star$ 
            that minimizes $\Delta C(i)$.\!
            \STATE Update the permutation 
            to $\perm$ on 
            $\{\bm{x}_1,  \dots, \bm{x}_{k}\}$
            by inserting $\bm{x}_k$ at the optimal position $i^\star$. 
            \STATE $\text{OTSD} \gets \text{OTSD} + \Delta C(i^\star)$.
        \ENDFOR        
        
        \STATE \textbf{Return} OTSD.
    \end{algorithmic}
\end{algorithm}

\paragraph{Normalized OTSD.}
To eliminate the influence of the problem dimensionality and ensure value consistency across different problems, we propose the \emph{normalized \ac{OTSD}}:
\begin{equation}
\label{eq:otsd-norm-def}
    \OTSD_\text{norm}(X_t) \coloneqq \frac{\OTSD(X_t)}{\Psi(d, t)},
\end{equation}
where $d$ denotes the dimensionality of the problem and $\Psi(d, t) \coloneqq 2\sqrt{5d}\left(\frac{3t}{2}\right)^{1-1/d}$ is the upper bound derived in Proposition~\ref{prop:otsd}. Unlike \ac{OTSD}, the normalized \ac{OTSD} is non-monotonic; a constant value indicates that the \ac{AF} does not change its behavior throughout the optimization, while higher and lower values suggest increased and decreased exploration, respectively.
We use the normalized \ac{OTSD} to aggregate results across different problems in Sec.~\ref{sec:experiment}.

\paragraph{Observation Entropy (OE).}
By treating observation points as samples from a random variable, we quantify the uniformity of the data distribution using empirical differential entropy. Higher entropy values indicate a more uniform spread of points, reflecting greater exploration. 
To estimate entropy without assuming an underlying distribution, we employ a non-parametric entropy estimator. Several methods exist, including histogram-based~\citep{gyorfi1987density}, kernel-density based~\citep{ahmad1976nonparametric}, and nearest-neighbor based~\citep{kozachenko1987statistical} approaches. 
Among them, the nearest-neighbor-based estimator, namely the \ac{KL} estimator, stands out for its sample efficiency and applicability in moderate-dimensional cases ($d\leq 50$), while other methods are very costly for $d\ge 10$.

We define \ac{OE} using the \ac{KL} estimator as follows:
\begin{equation}
\label{eq:kl-estimator}
\text{OE}(X_t) \coloneqq \frac{d}{t} \sum_{i=1}^{t} \log\left(\varepsilon_i^k\right) + \psi(t) - \psi(1) + \log V_d,
\end{equation}
where $\varepsilon_i^k$ denotes the distance between $\bm{x}_i$ and its $k$-th nearest neighbor, $V_d \coloneqq \pi^{d/2}/\Gamma(1+d/2)$ is the volume of the $d$-dimensional unit ball, $\Gamma(\cdot)$ denotes the Gamma function, and $\psi(t) \coloneqq \frac{\partial}{\partial t}\log \Gamma(t)$ is the digamma function. Following the recommendation of \citet{berrett2019efficient}, we set $k = \log(t)$ (using the natural logarithm), as increasing $k$ with $t$ improves the efficiency of the estimation.

\ac{OE} is non-monotonic and may take on either positive or negative values. 
An increasing \ac{OE} over \ac{BO} iterations indicates that the observed samples are more uniformly distributed, suggesting more exploration. 
Because of its non-monotonic nature, a sharp change in \ac{OE} signals that an \ac{AF} is altering its level of exploration over iterations. 
The computational complexity of \ac{OE} is $\mathcal{O}(dT^2)$ for $T$ observations in $d$ dimensions, or $\mathcal{O}(dTk)$ if distances are provided or $t\leq T$ gets updated sequentially. 
Algorithm~\ref{alg:kl-entropy} details the steps for computing \ac{OE}.

\begin{algorithm}
\caption{Kozachenko-Leonenko Entropy Estimation}
\label{alg:kl-entropy}
\begin{algorithmic}[1]
\REQUIRE Observation locations \(X_t = \{\bm{x}_1, \bm{x}_2, \dots, \bm{x}_t\}\)
\ENSURE Estimated observation entropy \(\text{OE}(X_t)\)
\STATE Compute pairwise distances: \(D(i,j) = \|\bm{x}_i - \bm{x}_j\|\) for all \(i,j \in \{1,\dots,t\}\) with \(i \neq j\).
\STATE For each point \(\bm{x}_i\), determine \(\varepsilon_i^k\) as the \(k\)-th smallest value in the set \(\{D(i,j) : j \neq i\}\).
\STATE Compute the volume of the unit ball in \(\mathbb{R}^d\):
$V_d = \frac{\pi^{d/2}}{\Gamma\bigl(1 + \tfrac{d}{2}\bigr)}$,
where \(\Gamma(\cdot)\) denotes the gamma function.
\STATE Evaluate the digamma functions \(\psi(t)\) and \(\psi(1)\), then compute the KL entropy estimator as given in Eq.~\eqref{eq:kl-estimator}.
\STATE \textbf{Return} \(\text{OE}(X_t)\).
\end{algorithmic}
\end{algorithm}

We illustrate \ac{OTSD}, normalized \ac{OTSD}, and \ac{OE} for various \acp{AF} on the 6-dimensional \texttt{Hartmann} function in Fig.~\ref{fig:exploration-hartmann}. The left panel shows the \ac{OTSD}, which increases monotonically as new observations are added. The center panel presents the normalized \ac{OTSD}, providing a clearer distinction among the different \acp{AF}. The right panel displays the \ac{OE} results. Since DM yields very low OE values (around $-134$), we exclude DM from the OE plot for better visualization. Overall, the figure demonstrates the explorative behavior of the various \acp{AF} and cross-validates the performance of \ac{OTSD} and \ac{OE}, as both yield consistent rankings.

The values of \ac{OTSD} and \ac{OE} are not directly comparable between different objective functions, as variations in input dimension, domain, and function landscape can significantly impact these quantities; neither quantity is invariant under a change of variables (for OE, this reflects that differential entropy is not invariant, unlike discrete entropy). 

\subsection{Exploration Bounds}
\begin{figure}[tb]
    \centering
    \includegraphics[width=\columnwidth]{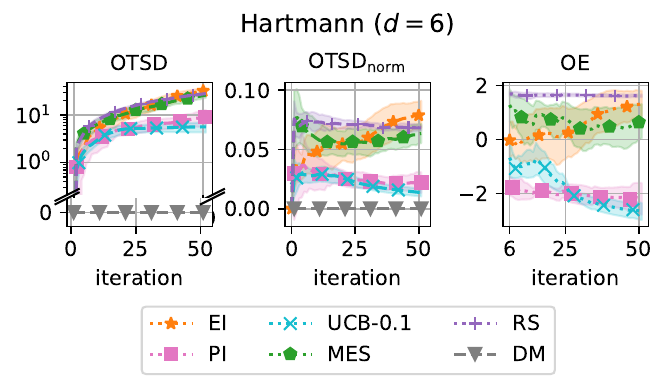}
    \caption{The exploration quantities \ac{OTSD} and \ac{OE} of RS, UCB ($\beta=0.1$), EI, PI, MES, and deterministic selection (DM) on the 6-dimensional \texttt{Hartmann} function. From left to right, these plots show \ac{OTSD}, normalized \ac{OTSD}, and \ac{OE}, respectively. DM values for OE (around $-134$) are hidden for better visualization. The shaded areas show the standard error of the mean.}
    \label{fig:exploration-hartmann}
\end{figure}
In Fig.~\ref{fig:exploration-hartmann}, we denote $\OTSD$ and $\OE$ as the statistical estimators based on the given input $X_t$. 
In this section, we establish bounds on their true values, denoted as $\OTSD^\ast(t)$ and $\OE^\ast(t)$ which depend only on $t$. 
The proofs for Propositions~\ref{prop:otsd} and~\ref{prop:oe} are detailed in Appendices~\ref{ssec:proof-otsd} and~\ref{ssec:proof-oe-bound}, respectively. 

\begin{proposition}[Upper and Lower Bounds for the True OTSD]
\label{prop:otsd}
Let $t$ denote the number of observations drawn from the unit cube $[0,1]^d$ in $d$-dimensional space ($d\geq 3$). Building on the bounds derived in \citep{bollobas1992travelling, balogh2024traveling}, the true observation traveling salesman distance $\OTSD^\ast(t)$ satisfies
\newcommand{\defeq}{\stackrel{\text{\tiny def}}{=}} 
\begin{equation}
    0 \leq \OTSD^\ast(t) \leq \Psi(d,t) \defeq 2\sqrt{5d}\left(\frac{3}{2}t\right)^{1 - 1/d}.
    \label{eq:bound_otsd}
\end{equation}
\end{proposition}
Note that the minimum traveling salesman distance is zero, which occurs when all points coincide.

\begin{proposition}[Upper Bound for the True OE]
\label{prop:oe}
For any probability distribution supported in the unit cube of dimension $d$, the uniform distribution achieves the maximum differential entropy with a zero value. 
Conversely, the differential entropy can be made arbitrarily negative, implying that there is no finite lower bound.
\end{proposition}
Typically, the sample set $X_t$ does not consist of independently and identically distributed points, as they are chosen adaptively via the Bayesian optimization process. Consequently, it is not necessarily the case that $\text{OE}(X_t) \le 0$, even when the samples lie within the unit hypercube. Further discussion is provided in Appendix~\ref{ssec:proof-oe-bias}.

\paragraph{Extension to Non-Euclidean Domains.}
In many practical problems, the input domain may be non-Euclidean. Examples include integers, ordinals, categoricals, protein sequences, strings, and graphs. Quantifying exploration on these non-Euclidean domains is particularly interesting. The \ac{OTSD} can still be applied in these settings provided a suitable metric is available since the insertion heuristic in Algorithm~\ref{alg:otsd} relies on the triangle inequality and thus only requires a metric space. 
However, the estimation of \ac{OE} becomes more challenging. In particular, the \ac{KL} estimator in Eq.~\eqref{eq:kl-estimator} is designed for estimating differential entropy in Euclidean spaces. Defining an appropriate notion of entropy and developing an estimator for such irregular spaces is a case-by-case problem that requires further investigation and is beyond the scope of this paper.

\section{Experiments}
\label{sec:experiment}
We evaluate \ac{OTSD} (including normalized \ac{OTSD}) and \ac{OE} on a wide range of synthetic and real-world benchmarks to tackle the following three research questions:
\begin{enumerate}[align=left, leftmargin=*]
    \item[RQ1.] Are the proposed quantities consistent with the literature, i.e., do \ac{OTSD} and \ac{OE} show higher exploration levels for \acp{AF} that are known to be more explorative?
    \item[RQ2.] How do \acp{AF}, whose exploration level has not yet been discussed, relate to others in terms of exploration?
    \item[RQ3.] What is the relationship between the level of exploration and optimization performance?
\end{enumerate}

\subsection{Experimental Setup}
\paragraph{Evaluation.} 
For each run of an \ac{AF} on a given benchmark, we record the locations $\bm{x}_i \in \mathcal{X}$ and their corresponding function values $y_{\bm{x}_i} \in \mathbb{R}$. 
From these observations, we compute the \ac{OTSD}, the \ac{OE}, and the performance, defined as the highest function value observed during the run. To aggregate OTSD results across different problems, we normalize OTSD using Eq.~\eqref{eq:otsd-norm-def} and then average these normalized values across the selected benchmarks. 
We give runtimes for \ac{OTSD} and \ac{OE} in Appendix~\ref{app:runtime}; \ac{OTSD} is fast to compute, requiring less than 200 ms while \ac{OE} needs $\approx 5$ s for 1,000 observations in a $20d$ space.
We also empirically validate Proposition~\ref{prop:otsd} in Appendix~\ref{sec:empirical-otsd}.

Since there is no straightforward method to normalize \ac{OE} and performance across problems with varying dimensions, we assess their relative rankings of competing methods. Specifically, we compute the mean \ac{OE} or performance for each method on each problem over ten repetitions, rank these mean values per problem, and then average the rankings across problems to obtain the \emph{mean relative ranking} (individual optimization performances for each benchmark are provided in Appendix~\ref{app:raw-optimization-performance}). Note that for \ac{OE}, the ranking is reversed so that more explorative methods receive a larger rank, ensuring consistency with the normalized \ac{OTSD} results.
Because the \ac{KL} estimator exhibits significant bias in high dimensions, we report \ac{OE} only for low-dimensional experiments ($d \leq 20$); for high-dimensional real-world experiments, we exclusively use \ac{OTSD}.

For better clarity, we only plot the mean values for normalized OTSD and the ranks.
In Appendix~\ref{app:error_bars}, we show the figures with the standard error of the mean.

\paragraph{Acquisition Functions.}
We study the following \acp{AF}: \acf{EI}, \acf{PI}, \acf{MES}, \acf{TS}, and \acf{UCB}.
We also include \acf{KG} in our comparison but only apply it to low-dimensional problems $(d\leq 10)$ due to its high computational cost.
We further compare with the popular \texttt{CMA-ES}~\citep{hansen2016cma}, which is an evolutionary method for gradient-free continuous non-convex optimization using the implementation by~\citet{hansen2019pycma}, version 3.3.0.
 
In addition to the simple \ac{BO} setup, we experiment with two common techniques for \ac{HDBO}: \acp{TR} and \ac{RAASP}, both introduced in the \texttt{TuRBO} algorithm~\citep{eriksson2019scalable}.
In \texttt{TuRBO}, \acp{TR}, \ac{TS}, and \ac{RAASP} are interwoven and not studied independently.
To allow assessing their individual effects, we therefore consider adaptations of these techniques.

Finally, we study the effect of batched evaluations.
In particular, we use the batched variants of the aforementioned \acp{AF} where available with batch sizes of $q=8$ and $q=32$.
While it is not uncommon to study the number of \emph{batch evaluations} to assess the effect of batching, we always study the number of function evaluations as we are interested in how batching changes exploration, i.e., we count one batch of size $q$ as $q$ individual function evaluations.

\paragraph{Benchmarks.}

We evaluate several \acp{AF} and variations thereof on nine benchmark problems, ranging from low-dimensional synthetic problems to noisy, high-dimensional simulations.
A summary of the benchmarks is given in Table~\ref{tab:benchmark_summary} in Appendix~\ref{app:details-experiments}.
The $2d$ Branin, $4d$ Levy, $6d$ Hartmann, and $8d$ Griewank problems are from~\citet{SFU_benchmarks}, the $8d$ Lasso-Diabetes and $180d$ Lasso-DNA problems from~\citet{vsehic2022lassobench}, the $60d$ Rover and $14d$ Robot Pushing problems from~\citet{wang2018batched}, and the $124d$ Mopta08 problem from~\citet{eriksson2021high}.

\subsection{Empirical Validation of OTSD and OE}

We begin by empirically validating that \ac{OTSD} and \ac{OE} effectively quantify exploration by applying them to methods that exhibit increasing levels of explorative behavior.
\begin{figure}[tb]
    \centering
    \includegraphics[width=1\linewidth]{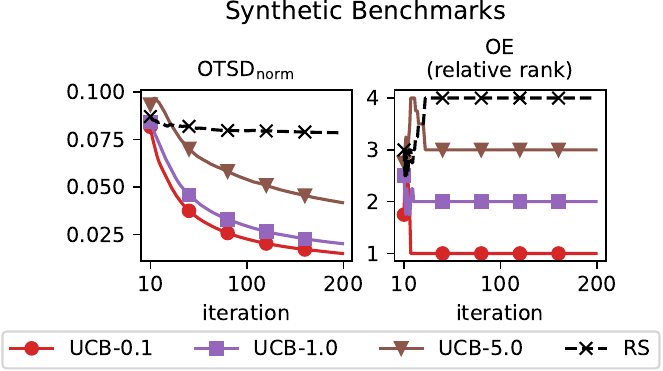}
    \caption{Normalized \ac{OTSD} and \ac{OE} rank averaged across all the synthetic benchmarks. A lower rank means lower exploration. We do not show the initial \acl{DoE} phase.}
    \label{fig:beta_ablation_ucb}
\end{figure}
Fig.~\ref{fig:beta_ablation_ucb} shows the normalized \ac{OTSD} (left) and the mean \ac{OE} ranks (right)
averaged over the synthetic benchmarks.
As expected, \ac{UCB} with a low $\beta=0.1$ (blue) achieves the lowest normalized \ac{OTSD} and mean \ac{OE} rank, followed by \ac{UCB} with a moderate $\beta=1$ (orange). 
\ac{UCB} with a high $\beta=5$ (green) achieves the highest  \ac{OTSD} and \ac{OE}.
The \ac{OTSD} curves going down indicate that, as expected, the \acp{AF} become more exploitative over time.
Appendix~\ref{app:empirical_validation_otsd_and_oe} shows the same result for real-world benchmarks; we also show the same analysis for \texttt{CMA-ES} with varying levels of $\sigma_0$. 

\subsection{Synthetic Benchmarks}

\begin{figure}[tb]
    \centering
    \includegraphics[width=\linewidth]{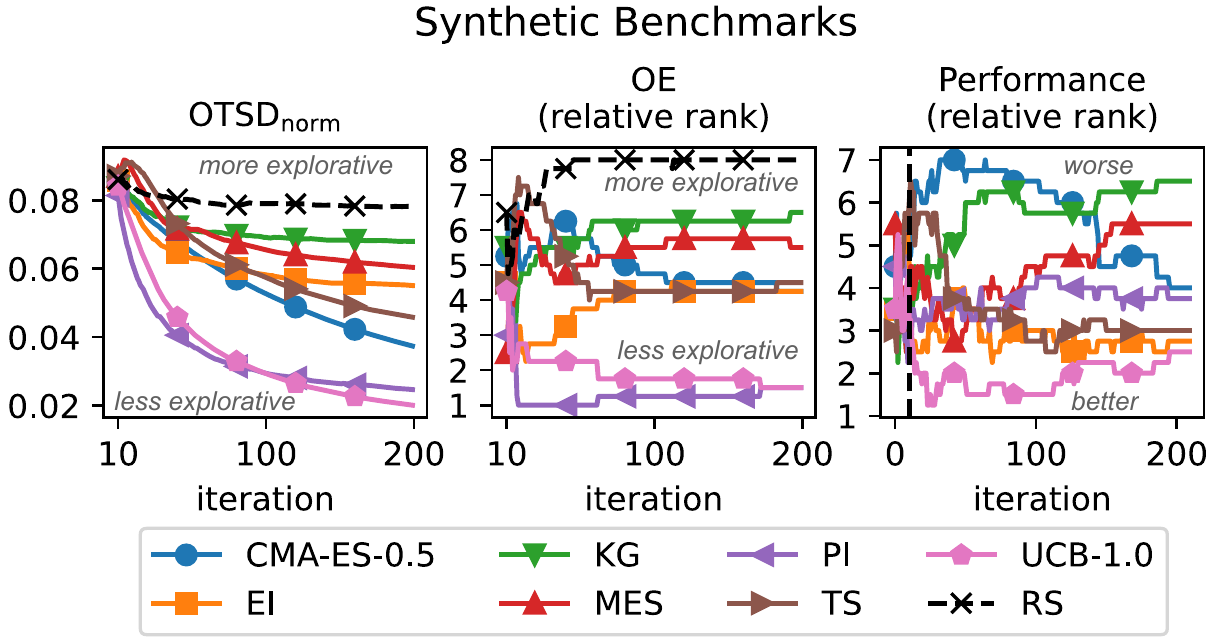}
    \caption{Normalized \ac{OTSD}, average rank for \ac{OE} and optimization performance on the synthetic benchmarks.}
    \label{fig:synthetic_benchmarks_basic_af_conf}
\end{figure}

Fig.~\ref{fig:synthetic_benchmarks_basic_af_conf} shows the performance of the various \ac{AF} (without \ac{RAASP} sampling and \acp{TR}), averaged over the four synthetic benchmarks.
We observe that at the start of the optimization, right after the \ac{DoE} phase, \ac{PI} is the least explorative \ac{AF} as it has low normalized \ac{OTSD} and the highest \ac{OE} relative rank of all \acp{AF}.
\ac{UCB}-1.0 is similarly unexplorative, starting only slightly more explorative than \ac{PI} and ending up overtaking \ac{PI} as the least explorative \ac{AF}.
In contrast, \ac{TS} starts as the most explorative \ac{AF} according to both \ac{OTSD} and \ac{OE}.
Eventually, \ac{MES}, \ac{KG}, and \texttt{CMA-ES} overtake \ac{TS} as the most explorative \ac{AF}.
\ac{EI} is on the same level of exploration as \ac{TS}. %
\ac{TS} and 
\ac{UCB}-1 shows the best relative rank optimization performance.

\begin{figure}[tb]
    \centering
    \includegraphics[width=\linewidth]{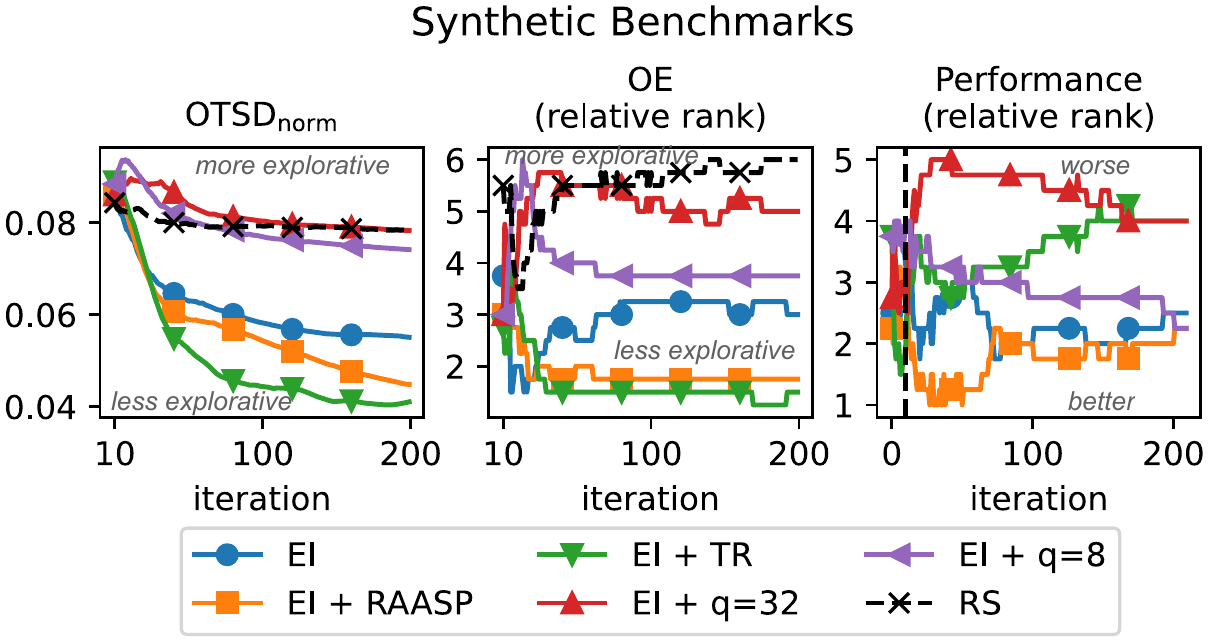}
    \caption{Normalized \ac{OTSD}, average ranks for \ac{OE} and optimization performance for \ac{EI} and its variations on the synthetic benchmarks.}
    \label{fig:synthetic_benchmarks_basic_ei}
\end{figure}

Next, we study the effect of batching, \ac{TR}, and \ac{RAASP} sampling on the level of exploration for \ac{EI} in 
Fig.~\ref{fig:synthetic_benchmarks_basic_ei}; see Appendix~\ref{app:other_af_variations} for the same setting on other \acp{AF}.
Batching increases exploration: \ac{EI} with a batch size of 32 (red) is the most explorative variant as indicated by both \ac{OTSD} and \ac{OE}.
The variant with the smaller batch size 8 (purple) is the next most explorative variant, followed by regular \ac{EI}.
\ac{RAASP} and \acp{TR}, on the other hand, lower the level of exploration.
Both \ac{RAASP} and \acp{TR} get a similarly low \ac{OE} rank, indicating that they are similarly unexplorative.
However, while \ac{RAASP} is the highest performing variant (orange), \acp{TR} reduce exploration on a similar level as \ac{RAASP}, but it is one of the worst variants, as shown in the right panel of Fig.~\ref{fig:synthetic_benchmarks_basic_ei}.
Batching also degrades optimization performance, as is expected when plotting against the number of function evaluations.

\begin{figure*}[t]
    \centering

    \includegraphics[width=.7\linewidth]{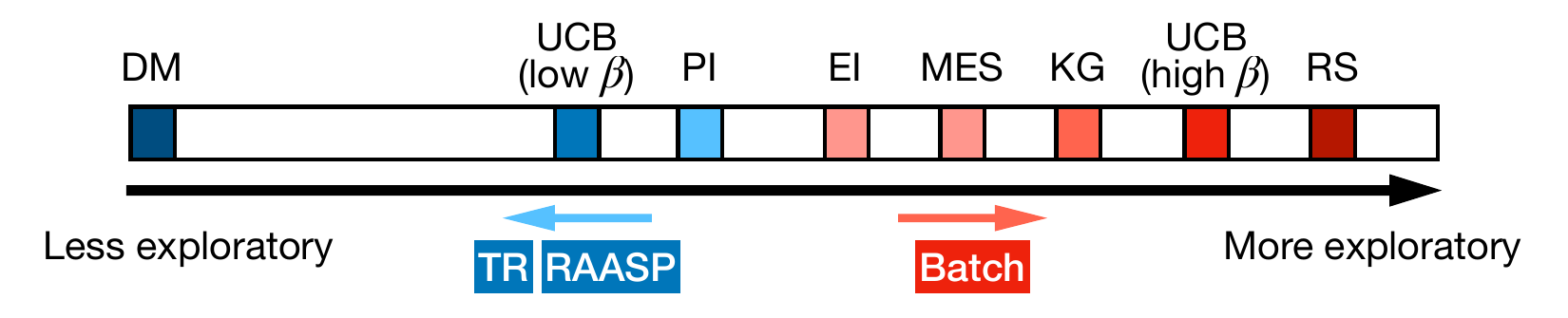}
    \caption{Empirical \ac{AF} (and their variants) exploration taxonomy based on the quantitative exploration methods \ac{OTSD} and \ac{OE}. It represents a general understanding and may differ in specific problems.}
    \label{fig:taxonomy}
\end{figure*}

\subsection{Real-World Benchmarks}

\begin{figure}[tb]
    \centering
    \includegraphics[width=.9\linewidth]{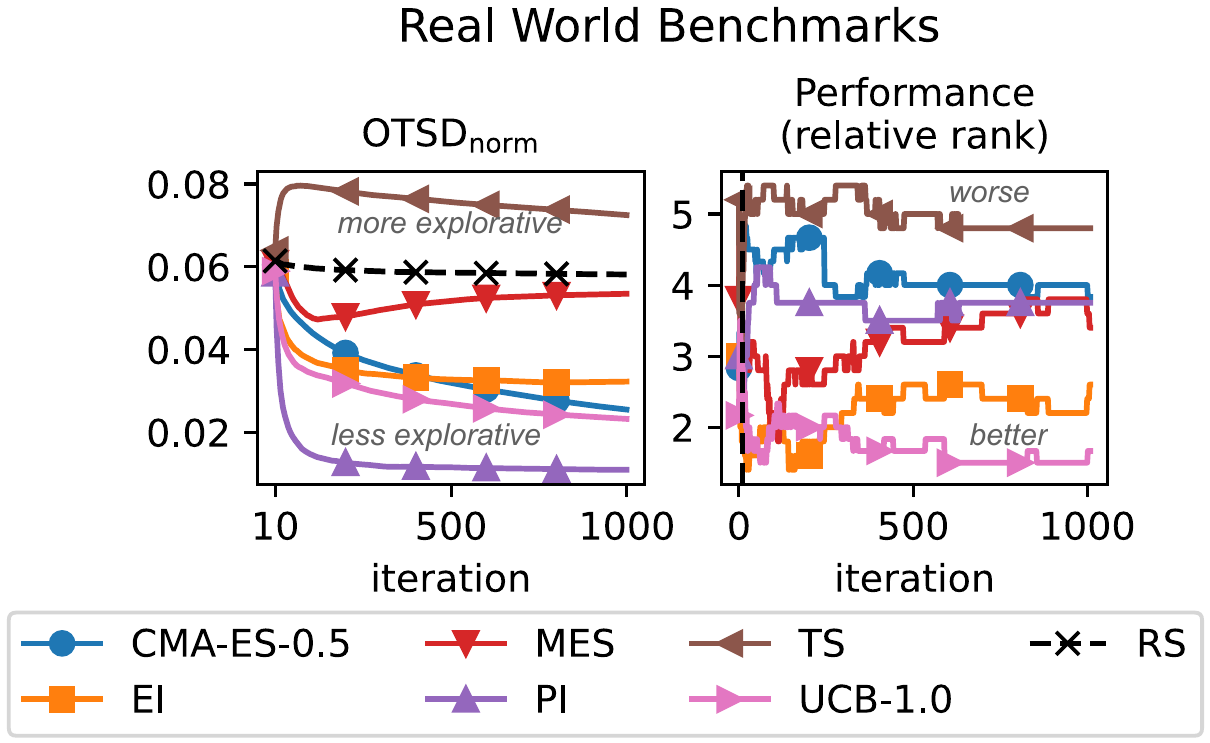}
    \caption{Normalized \ac{OTSD} and average optimization performance rank on the real-world benchmarks.}
    \label{fig:rw_benchmarks_basic_af_conf}
\end{figure}

Fig.~\ref{fig:rw_benchmarks_basic_af_conf} shows the performance of the various \ac{AF} (without \ac{RAASP} sampling and \acp{TR}) on the real-world benchmarks.
Since most real-world benchmarks are in high dimensions, we do not plot \ac{OE} for the aggregated results.
The exploration behavior in high dimensions is highly consistent, with minimal overlap between the OTSD curves. 
We hypothesize that \ac{EI} and \ac{UCB}-1.0 show the best optimization performance due to their balanced the \ac{EETO} as shown by the OTSD.
The overly-explorative \ac{TS} shows the worst performance, followed by the less-explorative \ac{PI}.
Compared to the other methods, \ac{MES} and \ac{TS} show a unique behavior: \ac{MES} initially becomes less explorative, indicated by a decreasing normalized \ac{OTSD}, but then reverses this trend and becomes more explorative.
\ac{TS} shows the opposite behavior.
Initially, it strives for pure exploration and is even more explorative than \ac{RS}.
We explain this behavior with \ac{TS} choosing areas where the surrogate exhibits high posterior variance, choosing points distant to previous observations and hence yielding high \ac{OTSD}.
Later on, as the posterior variance of the surrogate decreases, \ac{TS} becomes less explorative.

We also study the effect of \acp{TR}, \ac{RAASP}, and batching on the real-world benchmarks.
The results are similar to the synthetic case, so we save them for Appendix~\ref{app:trs-raasp-batching-real-world}.

\paragraph{Impact of the Dimensionality.}
In addition to distinguishing between synthetic and real-world scenarios, we examine \ac{OTSD} and \ac{OE} across low- and high-dimensional regimes. Two key differences emerge: first, \ac{TS} is considerably more explorative in high-dimensional benchmarks; second, while \ac{EI} is more explorative than other methods in low dimensions, it becomes less so in high dimensions. One potential explanation is that in high-dimensional spaces, the \ac{GP} may struggle to accurately learn the objective function, further encouraging explorative sampling, which significantly impacts \ac{TS}. Moreover, lengthscales are underestimated in higher dimensions, causing \ac{EI} to make more conservative predictions than in low-dimensional settings. Due to space limitations, we present their synthetic problem results in Appendix~\ref{app:impact-of-dimensionality}.

\subsection{Exploration Taxonomy}
Our empirical results on synthetic and real-world benchmarks confirm the partial ordering $\text{KG} \succeq \text{EI} \succeq \text{PI}$, which was common knowledge within the community as discussed in Sec.~\ref{sec:analysis_of_tribal_knowledge}. 
Furthermore, our findings reveal that \ac{KG} is slightly more explorative than \ac{MES}, addressing a missing link in previous studies. \ac{TS} is challenging to classify because its behavior varies dramatically between low and high dimensions. In low dimensions, \ac{TS} performs similarly to \ac{EI}; while it becomes overly explorative in high dimensions -- surpassing even \ac{RS} in terms of normalized \ac{OTSD}. 
Expanding the exploration taxonomy, we empirically demonstrate that techniques such as \acfp{TR} and \ac{RAASP} consistently reduce exploration, whereas batching tends to increase exploration. Finally, our findings indicate that the best-performing methods tend not to exhibit extreme exploration quantities; their \ac{OTSD} and \ac{OE} values are neither excessively high nor excessively low compared to other approaches. However, they are often on the less explorative side of the spectrum, as demonstrated by the decrease in OTSD during the optimization iterations.
To summarize our results, we present a revised empirical \ac{AF} exploration taxonomy in Fig.~\ref{fig:taxonomy} based on our new \ac{OTSD} and \ac{OE} quantities.

\section{Conclusion}

In this work, we introduce two novel methods and their theoretical bounds to quantify the magnitude of exploration in various acquisition functions. We conduct extensive experiments on synthetic and real-world benchmarks, spanning low- and high-dimensional settings, to evaluate commonly used acquisition functions and their variants. Our empirical findings demonstrate that \ac{OTSD} and \ac{OE} effectively capture the level of exploration exhibited by these acquisition functions. Finally, we present the first empirical taxonomy of acquisition function exploration based on these quantification methods. 
These results offer valuable insights for designing new acquisition functions, constructing acquisition function portfolios, or controlling the optimization.
For instance, an \ac{AF} having higher \ac{OTSD} than a \acl{RS} (see \ac{TS} in Fig.~\ref{fig:rw_benchmarks_basic_af_conf}) can serve as a warning sign that this \ac{AF} is too explorative for the problem at hand.
In such situations, one could either switch to more local \ac{AF} according to the taxonomy or combine the \ac{AF} with \ac{RAASP} sampling.
Similarly, if an \ac{AF} approaches the \ac{OTSD} of a \acl{RS} after being less explorative at the beginning of the optimization (see \ac{MES} in Fig.~\ref{fig:rw_benchmarks_basic_af_conf}), it could be an early-stopping indicator where the \ac{AF} exhaustively visited all local minima and the optimization can come to an end.

\paragraph{Limitations and Future Work.}
While our results are obtained by extensive empirical experimentation, they do not consider the effect of \ac{GP} hyperparameters and hyperpriors on the behavior of \acp{AF} and are limited to continuous domains.
In the future, we will expand the taxonomy to include additional \acp{AF}, for instance, other information-theoretic \acp{AF}~\citep{hennig2012entropy, hernandez2014predictive, hvarfner2022joint, cheng2025unified}, and study the effect of kernel and likelihood functions and their hyperparameters on the behavior of \acp{AF} in \acl{BO}.
Furthermore, we will study if \ac{OTSD} and \ac{OE} can be extended to other domains, such as non-Euclidean spaces.

\begin{acknowledgements}
    This project was partly supported by the Wallenberg AI, Autonomous Systems, and Software program (WASP) funded by the Knut and Alice Wallenberg Foundation.
    The computations were enabled by resources provided by the National Academic Infrastructure for Supercomputing in Sweden (NAISS), partially funded by the Swedish Research Council through grant agreement no. 2022-06725
\end{acknowledgements}

\bibliography{main}

\newpage

\onecolumn

\title{Appendix\\(Supplementary Material)}
\maketitle

\appendix

\section{Proofs}
\label{sec:proof}

\subsection{OTSD Upper Bound}
\label{ssec:proof-otsd}
\begin{proof}
Our proof follows from \citet[Theorem 1.3]{balogh2024traveling}. In that work, the authors show that the $d$-norm length of a Hamiltonian cycle with $t$ nodes in a $d$-dimensional unit cube, denoted by 
\begin{equation}
s^\text{HC}_d(t) \coloneqq \left(\sum_{i=1}^t \sqrt{\mathrm{e}_i^d}\right)^{1/d},
\end{equation}
satisfies
\begin{equation}
s^\text{HC}_d(t) \leq 3\sqrt{5}\left(\frac{2}{3}\right)^{1/d}\sqrt{d},
\end{equation}
where
\begin{equation}
\mathrm{e}_i \coloneqq \lVert \bm{x}_{\perm(i)} - \bm{x}_{\perm(i)+1}\rVert_2
\end{equation}
represents the Euclidean distance between successive nodes.

The true total distance with $t$ nodes, denoted by $\OTSD^\ast(t)$, is given by
\begin{equation}
\OTSD^\ast(t) \coloneqq \sum_{i=1}^t \mathrm{e}_i.
\end{equation}
Applying Hölder's inequality,
\begin{equation}
\lVert\bm{e}\rVert_1 \leq \lVert\bm{e}\rVert_d \cdot \lVert\bm{1}\rVert_{(1-1/d)^{-1}},
\end{equation}
yields
\begin{equation}
\OTSD^\ast(t) = \sum_{i=1}^t \mathrm{e}_i \leq s^\text{HC}_d(t) \cdot \left(\sum_{i=1}^t 1\right)^{1-1/d}.
\end{equation}
Since $\sum_{i=1}^t 1 = t$, it follows that
\begin{equation}
\OTSD^\ast \leq 3\sqrt{5}\left(\frac{2}{3}\right)^{1/d}t^{1-1/d}\sqrt{d} = 2\sqrt{5d}\left(\frac{3}{2}t\right)^{1 - 1/d}.
\end{equation}
\end{proof}

\subsection{OE Upper Bound}
\label{ssec:proof-oe-bound}
We show that the distribution with maximum differential entropy in the unit cube $[0, 1]^d$ is the uniform distribution.

\begin{proof}
We wish to determine the probability density function $p(\bm{x})$, for $\bm{x}\in[0,1]^d$, that maximizes the differential entropy
\begin{equation}
    H[p] = -\int_{[0,1]^d} p(\bm{x})\ln p(\bm{x})\,\mathrm{d}\bm{x},
\end{equation}
subject to the normalization constraint
\begin{equation}
    \int_{[0,1]^d} p(\bm{x})\,\mathrm{d}\bm{x} = 1.
\end{equation}

To enforce this constraint, we introduce a Lagrange multiplier $\lambda$ and consider the augmented functional
\begin{equation}
    \mathcal{L}[p] = -\int_{[0,1]^d} p(\bm{x})\ln p(\bm{x})\,\mathrm{d}\bm{x} + \lambda\left(\int_{[0,1]^d} p(\bm{x})\,\mathrm{d}\bm{x} - 1\right).
\end{equation}

We then compute the first variation $\delta \mathcal{L}$ with respect to an arbitrary variation $\delta p(\bm{x})$, yielding
\begin{equation}
    \delta \mathcal{L} = -\int_{[0,1]^d} \delta p(\bm{x})\Bigl[\ln p(\bm{x}) + 1\Bigr]\,\mathrm{d}\bm{x} + \lambda\int_{[0,1]^d} \delta p(\bm{x})\,\mathrm{d}\bm{x}.
\end{equation}

For $\delta \mathcal{L}$ to vanish for all admissible variations $\delta p(\bm{x})$, the integrand must be zero:
\begin{equation}
    -\ln p(\bm{x}) - 1 + \lambda = 0 \quad \text{for all } \bm{x}\in[0,1]^d.
\end{equation}

Solving for $\ln p(\bm{x})$, we obtain
\begin{equation}
    \ln p(\bm{x}) = \lambda - 1,
\end{equation}
which implies that
\begin{equation}
    p(\bm{x}) = e^{\lambda - 1}.
\end{equation}

Since $p(\bm{x})$ is constant over $[0,1]^d$, we can determine the constant by imposing the normalization condition:
\begin{equation}
    \int_{[0,1]^d} p(\bm{x})\,\mathrm{d}\bm{x} = e^{\lambda - 1} \cdot 1 = 1.
\end{equation}

Thus, 
\begin{equation}
    e^{\lambda - 1} = 1 \quad \Longrightarrow \quad \lambda - 1 = 0 \quad \Longrightarrow \quad \lambda = 1,
\end{equation}
yields
\begin{equation}
    p(\bm{x}) = 1 \quad \text{for all } \bm{x}\in[0,1]^d.
\end{equation}

This completes the proof that the maximum entropy distribution on the $d$-dimensional unit cube is indeed the uniform distribution. Therefore, 
\begin{equation}
    \OE^\ast(t) \leq H[p_\text{unif}] = 0.
\end{equation}
\end{proof}

\subsection{KL Estimation Consistency and Bias}
\label{ssec:proof-oe-bias}

Many studies have examined the estimation bias of the Kozachenko-Leonenko estimator (see Eq.~\eqref{eq:kl-estimator}). The original work by \citet{kozachenko1987statistical} established the consistency of the estimator under mild conditions when $k=1$. Furthermore, \citet{pal2010estimation} demonstrated both the consistency and the convergence rate of the nearest-neighbor-based estimator for R\'enyi entropies under the assumption that the entropy support is bounded. In addition, \citet{delattre2017kozachenko} extended these results to non-compactly supported densities, providing an upper bound for the bias of order $\mathcal{O}(t^{-2/d})$. A more recent study~\citep{devroye2021consistency} presents a consistency result for the KL estimator even when the density function is not smooth. However, in our study the sample points in $X_t$ are generally not independent and identically distributed, as they are collected via a Bayesian optimization process. Consequently, we believe that the practical performance of OE does not align with the findings of these previous studies.

\section{Details on the Experimental Setup}
\label{app:details-experiments}

We run each \ac{AF} in a basic \ac{BO} setup where we first initialize the \ac{GP} with ten observations that we sample uniformly at random from $\mathcal{X}$ and then start the \ac{BO} loop for 200 iterations for synthetic and 1000 iterations for real-world problems.
Similarly, we run \texttt{CMA-ES} with a population size of 5 with different initial step sizes $\sigma_0\in\{0.05, 0.1, 0.5\}$.
\ac{UCB} allows one to specify an exploration parameter $\beta$, which we set to $\beta\in\{0.1, 1, 5\}$.

To evaluate the \acp{AF}, we use the default \texttt{SingleTaskGP} \ac{GP} model provided by \texttt{BoTorch} version \texttt{0.12.0}~\citep{balandat2020botorch} as well as \texttt{BoTorch}'s provided methods to fit the \ac{GP} and maximize the \ac{AF}.
In the case of \ac{TS}, we sample from the \ac{GP} posterior using pathwise conditioning~\citep{wilson2021pathwise} and maximize the posterior sample using \texttt{BoTorch}'s \texttt{optimize\_posterior\_samples} function with 1024 initial random samples and 20 restarts of the \ac{GD} optimizer.

For \ac{RAASP} sampling, we rely on the \texttt{sample\_around\_best} parameter for \texttt{BoTorch}'s \ac{AF} optimizer that, with a probability of $\min\left(1,\frac{20}{d}\right)$, substitutes a parameter's current best configuration with a value sampled from a truncated Gaussian, centered on the incumbent observation.
For \acp{TR}, we follow \texttt{TuRBO}'s specification for finding the \ac{TR} bounds.
We then configure the \ac{AF} maximizer to respect these bounds, similar to~\citet{TuRBO_BoTorch} but dropping the sparse perturbations. 

Table~\ref{tab:benchmark_summary} summarizes the benchmarks used in this work.

\begin{table}[tbh]
    \centering
    \begin{tabular}{lrcc}
        \toprule
         Name & $d$ & Noise & Synth. \\
         \midrule
         Branin & 2 & \xmark & \cmark \\
         Levy & 4 & \xmark & \cmark \\
         Hartmann & 6 & \xmark & \cmark \\
         Griewank & 8 & \xmark & \cmark\\
         Lasso-Diabetes & 8 & \xmark & \xmark\\
         Robot Pushing & 14 & \cmark & \xmark\\
         Rover & 60 & \xmark & \xmark\\
         Mopta08 & 124 & \xmark & \xmark\\
         Lasso-DNA & 180 & \xmark & \xmark\\
         \bottomrule
    \end{tabular}
    \caption{Benchmark summary.}
    \label{tab:benchmark_summary}
\end{table}

\section{Empirical Verification of the OTSD Bound}
\label{sec:empirical-otsd}

To verify the upper bound stated in Proposition~\ref{prop:otsd}, we take advantage of the normalized OTSD defined in Eq.~\eqref{eq:otsd-norm-def} which makes the OTSD independent from the dimensionality $d$. If the approximation error to compute OTSD is ignored, the theoretical bound implies that $\OTSD_{\text{norm}}(t)$ should remain below 1 for all $t$ (or with approximation error considered, $\OTSD_{\text{norm}}(t)$ should remain below 2 for all $t$).

\begin{figure}[htb]
    \centering
    \includegraphics[width=0.5\linewidth]{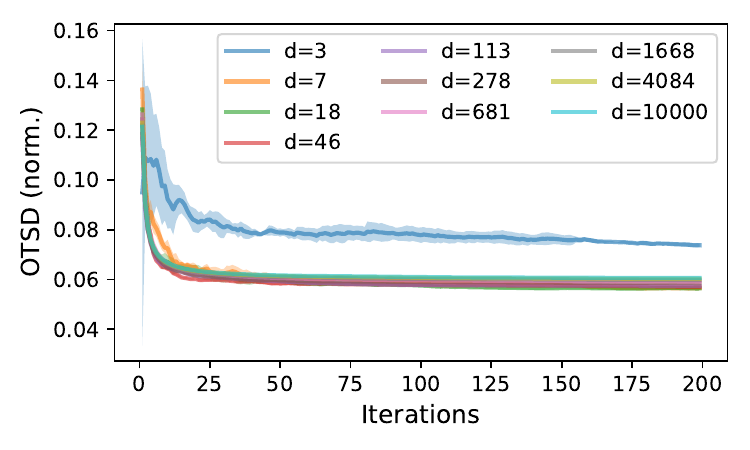}
    \caption{Normalized OTSD of Random Search for varying problem dimensions.}
    \label{fig:rs_otsd}
\end{figure}

We show $\OTSD_{\text{norm}}$ for Random Search (\ac{RS}) across various dimensionalities of the problem in Fig.~\ref{fig:rs_otsd}.
As the figure illustrates, all values remain well below the theoretical threshold of 1. Given that \ac{RS} represents a highly explorative scenario, this result empirically corroborates the bound in Proposition~\ref{prop:otsd}. Moreover, we observe that for higher-dimensional problems, the normalized OTSD converges to a specific constant, whereas for lower-dimensional problems, the convergence is less pronounced. 
This suggests that while the bound is reliable in high dimensions, it may not be as tight in low dimensions.

\section{Additional Experiments}
\label{app:benchmark-overview}

\subsection{Empirical Validation of OTSD and OE}
\label{app:empirical_validation_otsd_and_oe}

In this section, we study the \ac{OTSD} for varying $\beta$-values for \ac{UCB} in real-world settings and for \texttt{CMA-ES} with different step sizes, further supporting the adequacy of \ac{OTSD} and \ac{OE} for quantifying exploration.

\begin{figure}[H]
    \centering
    \begin{subfigure}{0.48\linewidth}
    \includegraphics[width=\linewidth]{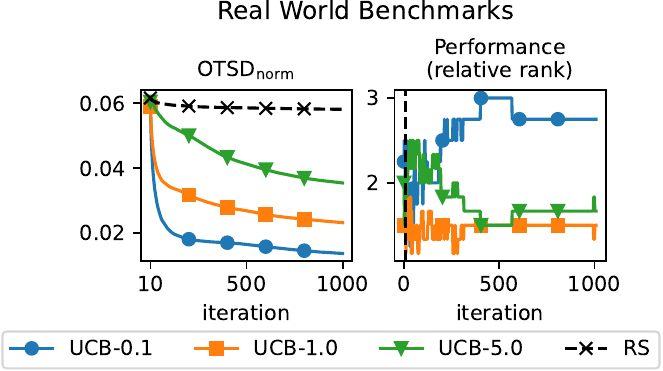}
    \caption{Without error bars.}
    \end{subfigure}
    \begin{subfigure}{0.48\linewidth}
    \includegraphics[width=\linewidth]{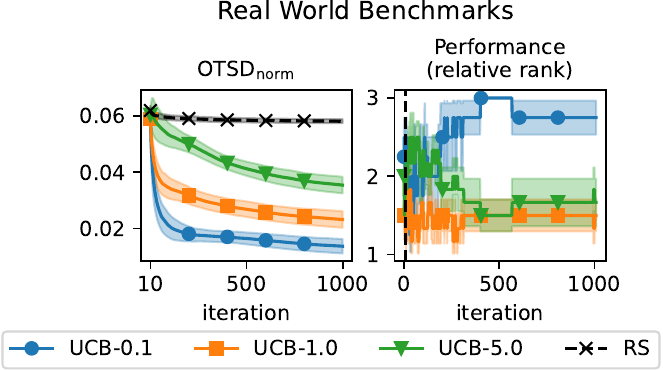}
    \caption{With error bars.}
    \end{subfigure}
    \caption{Normalized OTSD and mean ranks of the empirical performance for \ac{UCB} with varying $\beta$-parameters on the real-world benchmarks.}
    \label{fig:abl_ucb_rw}
\end{figure}

\begin{figure}[H]
    \centering
    \begin{subfigure}{0.48\linewidth}
    \includegraphics[width=\linewidth]{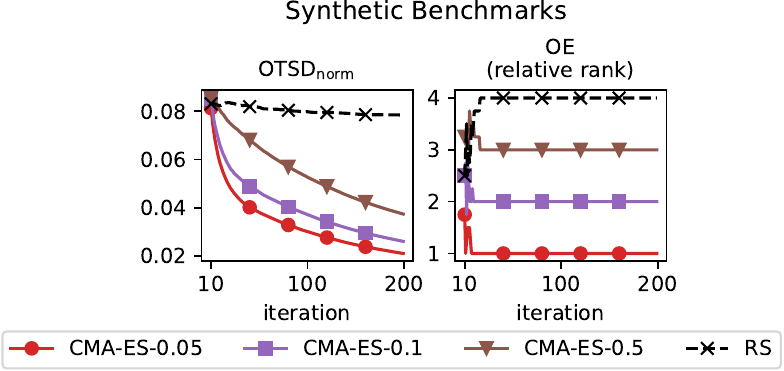}
    \caption{Without error bars.}
    \end{subfigure}
    \begin{subfigure}{0.48\linewidth}
    \includegraphics[width=\linewidth]{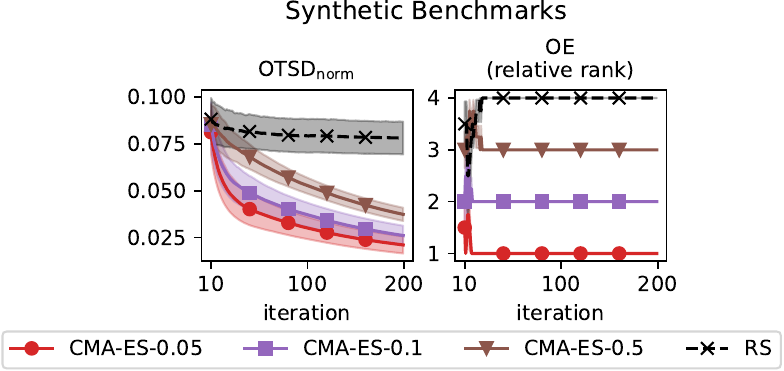}
    \caption{With error bars.}
    \end{subfigure}
    \caption{Normalized OTSD and mean ranks of the empirical performance for \texttt{CMA-ES} with varying $\sigma_0$-parameters on the synthetic benchmarks.}
    \label{fig:abl_cmaes_synthetic}
\end{figure}

\begin{figure}[H]
    \centering
    \begin{subfigure}{0.48\linewidth}
    \includegraphics[width=\linewidth]{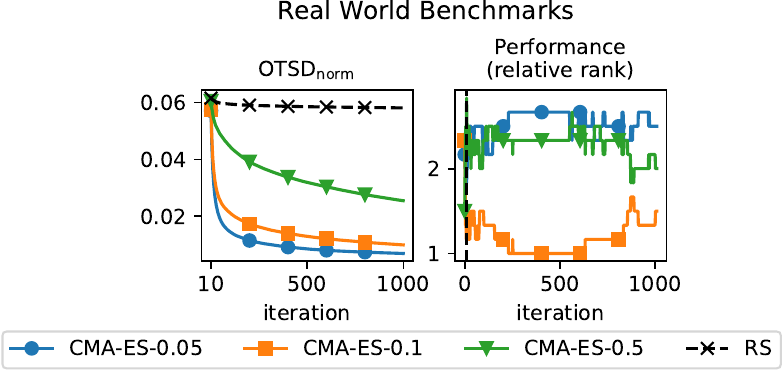}
    \caption{Without error bars.}
    \end{subfigure}
    \begin{subfigure}{0.48\linewidth}
    \includegraphics[width=\linewidth]{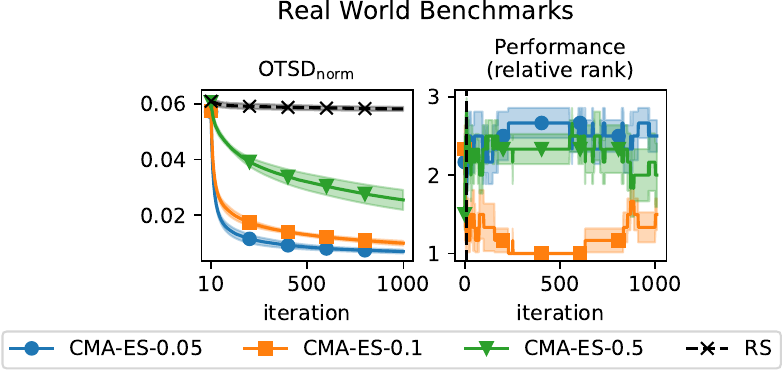}
    \caption{With error bars.}
    \end{subfigure}
    \caption{Normalized OTSD and mean ranks of the empirical performance for\texttt{CMA-ES} with varying $\sigma_0$-parameters on the real-world benchmarks.}
    \label{fig:abl_cmaes_rw}
\end{figure}

\subsection{Error Bars for Main Text Figures}
\label{app:error_bars}

We report the figures from the main text with error bars indicating the standard error of the mean.

\begin{figure}[H]
    \centering
    \includegraphics[width=0.65\linewidth]{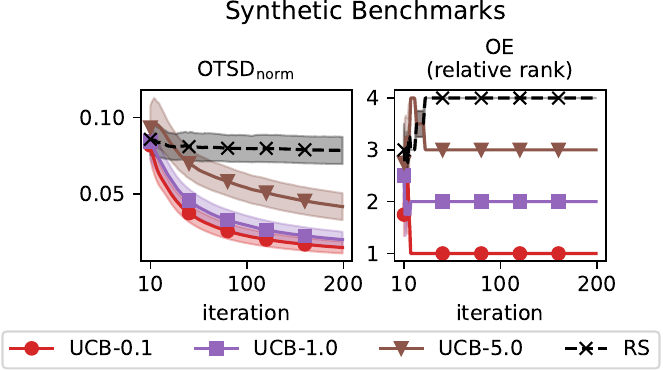}
    \caption{Normalized \ac{OTSD} and \ac{OE} rank averaged across all the synthetic benchmarks. A lower rank means lower exploration. We do not show the initial \acl{DoE} phase.}
\end{figure}

\begin{figure}[H]
    \centering
    \includegraphics[width=.65\linewidth]{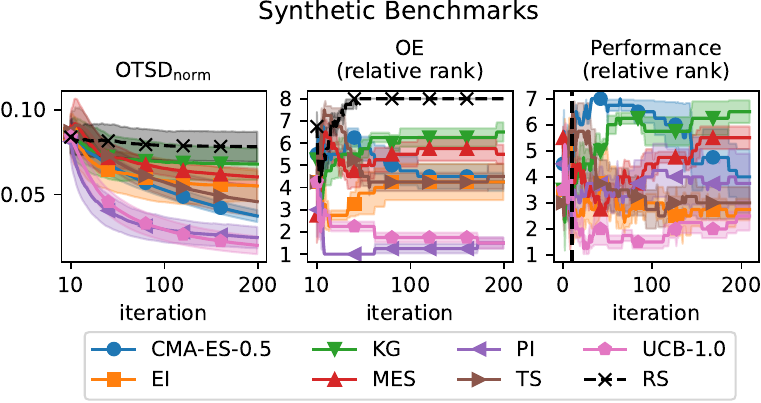}
    \caption{Normalized \ac{OTSD}, average rank for \ac{OE} and optimization performance on the synthetic benchmarks.}
\end{figure}

\begin{figure}[tb]
    \centering
    \includegraphics[width=.65\linewidth]{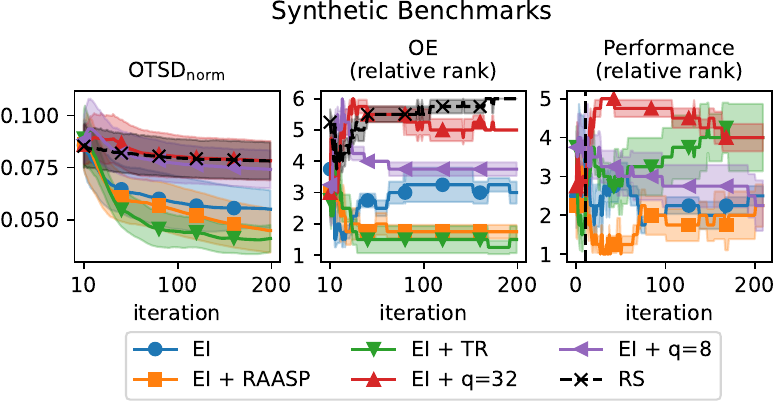}
    \caption{Normalized \ac{OTSD}, average ranks for \ac{OE} and optimization performance for \ac{EI} and its variations on the synthetic benchmarks.}
\end{figure}

\begin{figure}[tb]
    \centering
    \includegraphics[width=.65\linewidth]{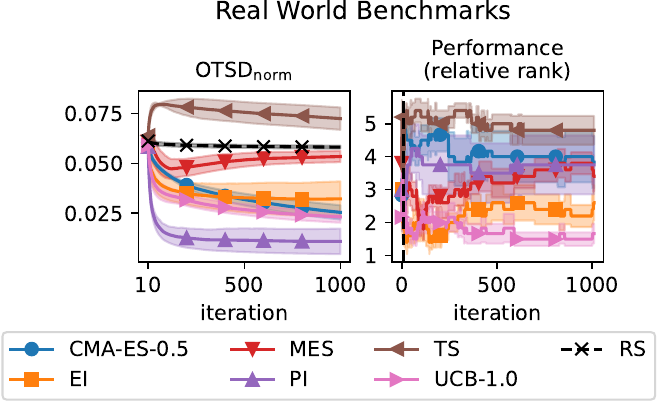}
    \caption{Normalized \ac{OTSD} and average optimization performance rank on the real-world benchmarks.}
\end{figure}

\subsection{TRs, RAASP, and batching on the Real-World Benchmarks}
\label{app:trs-raasp-batching-real-world}

\begin{figure}[H]
    \centering
    \begin{subfigure}{0.48\linewidth}
    \includegraphics[width=\linewidth]{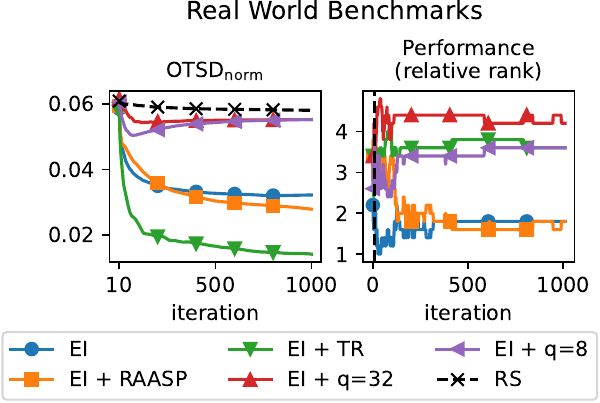}
    \caption{Without error bars.}
    \end{subfigure}
    \begin{subfigure}{0.48\linewidth}
    \includegraphics[width=\linewidth]{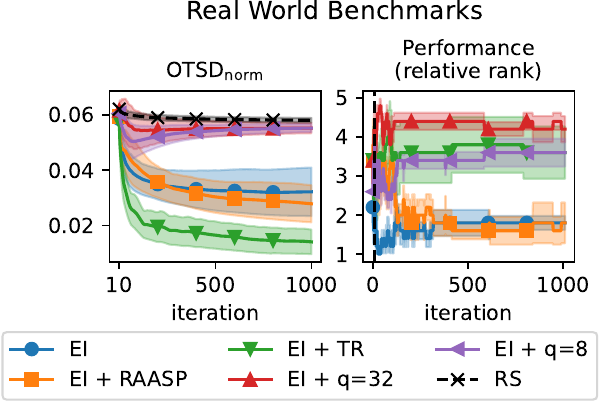}
    \caption{With error bars.}
    \end{subfigure}
    \caption{Normalized \ac{OTSD} and average optimization performance rank on the real-world benchmarks for \ac{EI} and its variations. \acp{TR} and \ac{RAASP} sampling promote exploitation.}
    \label{fig:rw_benchmarks_basic_ei}
\end{figure}

Fig.~\ref{fig:rw_benchmarks_basic_ei} zooms in on the effect of batching, \ac{TR}, and \ac{RAASP} sampling.
Here, the behavior is similar to the synthetic benchmarks: batching increases exploration and larger batch sizes lead to more exploration while \ac{RAASP} sampling and \acp{TR} reduce exploration.
In particular, the level of exploitation introduced by the \acp{TR} dominates all other methods. 
Compared to the optimization performance, both over-exploration and over-exploitation get punished: the most and least explorative methods (`EI + TR' and `EI + q=32') show the worst empirical performance as indicated by the high average rank of the purple and blue curves in the right panel of Fig.~\ref{fig:rw_benchmarks_basic_ei}.

\subsection{Impact of the Dimensionality}
\label{app:impact-of-dimensionality}
\begin{figure}[H]
    \centering
    \begin{subfigure}{0.48\linewidth}
    \includegraphics[width=\linewidth]{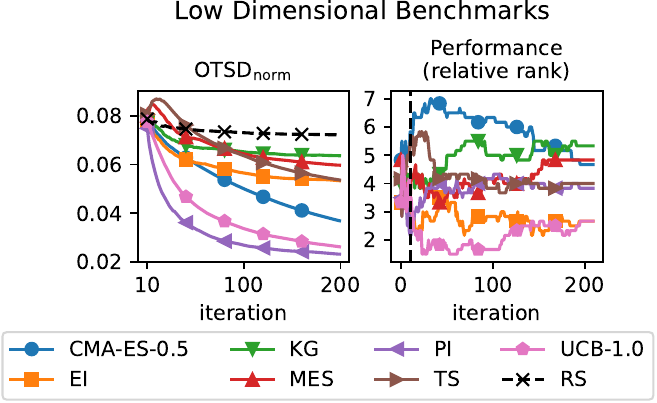}
    \caption{Without error bars.}
    \end{subfigure}
    \begin{subfigure}{0.48\linewidth}
    \includegraphics[width=\linewidth]{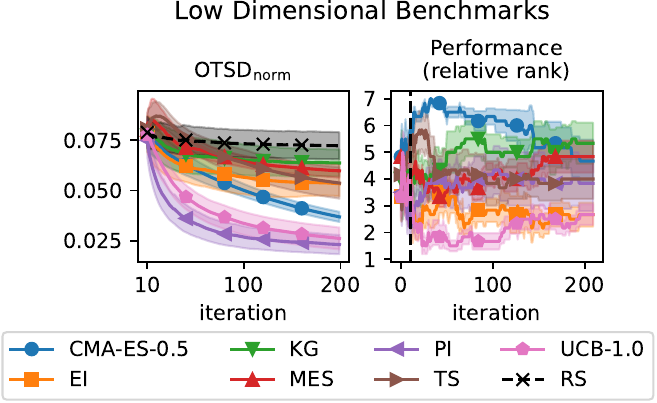}
    \caption{With error bars.}
    \end{subfigure}
    \caption{Normalized \ac{OTSD} and optimization performance ranks on the low-dimensional problems. }
    \label{fig:lowdim_benchmarks_basic_af_conf}
\end{figure}

\begin{figure}[H]
    \centering
    \begin{subfigure}{0.48\linewidth}
    \includegraphics[width=\linewidth]{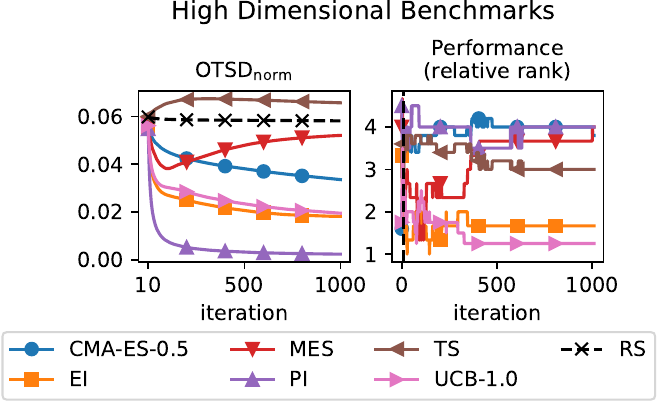}
    \caption{Without error bars.}
    \end{subfigure}
    \begin{subfigure}{0.48\linewidth}
    \includegraphics[width=\linewidth]{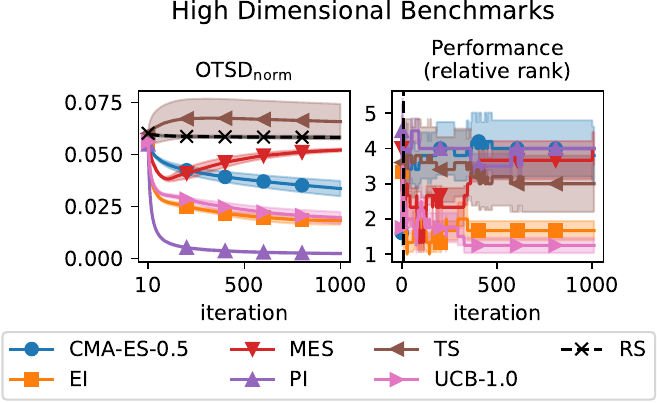}
    \caption{With error bars.}
    \end{subfigure}
    \caption{Normalized \ac{OTSD} and optimization performance rank on the high-dimensional problems.}
    \label{fig:highdim_benchmarks_basic_af_conf}
\end{figure}

Next, we study how the dimensionality of problems affects exploration.
To this end, we compare low-dimensional ($d\leq 20$, Fig.~\ref{fig:lowdim_benchmarks_basic_af_conf}) and high-dimensional problems ($d> 20$, Fig.~\ref{fig:highdim_benchmarks_basic_af_conf}), unveiling significant differences between low- and high-dimensional regimes.
While \ac{TS} is eventually overtaken by \ac{MES} in terms of exploration, it is by far the most explorative \ac{AF} on high-dimensional problems.
This is arguably due to the vast regions of high posterior uncertainty in high-dimensional spaces that allow diverse posterior samples.
Similarly, \ac{MES} is more explorative than other \acp{AF} (except for \ac{TS}) in high-dimensional than low-dimensional spaces.
Conversely, \ac{EI} is considerably more explorative than \ac{UCB}-1 in low-dimensional but not high-dimensional spaces.
Arguably, the fast collapse of posterior uncertainty in low-dimensional spaces affects \ac{UCB}-1 more than \ac{EI}, making \ac{UCB} almost as exploitative as \ac{PI}.
In both regimes, \ac{PI} is most exploitative while showing mediocre to bad optimization performance.
Brought together that the over-explorative \ac{TS} also shows subpar optimization performance, we reinforce our conclusion that a balanced \ac{EETO} is crucial for successful black-box optimizers.

\subsection{Optimizer Variations on Different Acquisition Functions}
\label{app:other_af_variations}

Here, we study the effect of \ac{RAASP} sampling, \acp{TR}, and batching on all \acp{AF} used in Section~\ref{sec:experiment}.

\subsubsection{Probability of Improvement}

For \ac{PI}, we did not study batching as it is not implemented in \texttt{BoTorch}, so we only focus on \acp{TR} and \ac{RAASP} sampling.

\begin{figure}[H]
    \centering
    \begin{subfigure}{0.48\linewidth}
    \includegraphics[width=\linewidth]{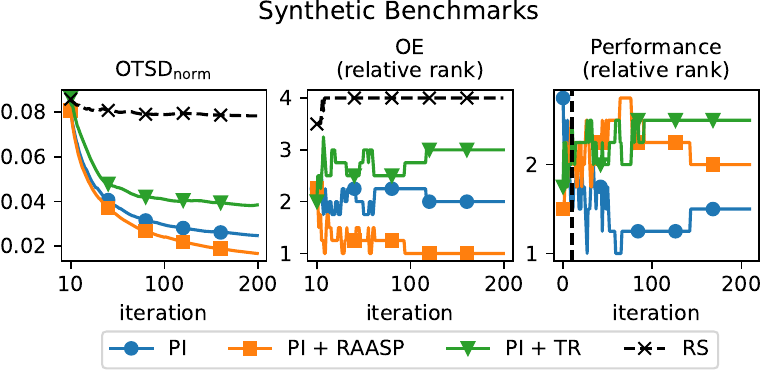}
    \caption{Without error bars.}
    \end{subfigure}
    \begin{subfigure}{0.48\linewidth}
    \includegraphics[width=\linewidth]{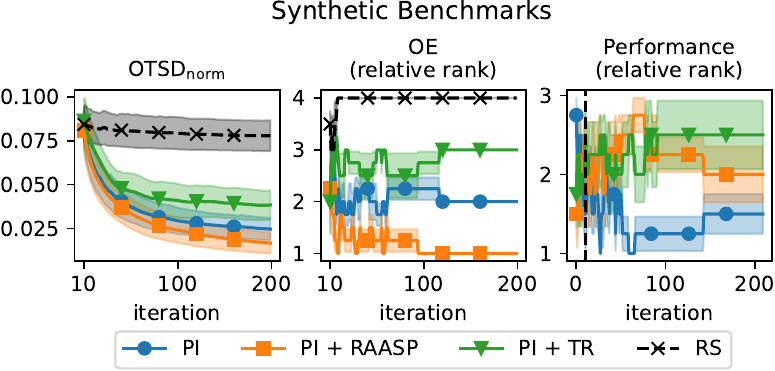}
    \caption{With error bars.}
    \end{subfigure}
    \caption{Effect of \acp{TR} and \ac{RAASP} sampling on \ac{PI} in the context of synthetic benchmarks: Both methods reduce the level of exploration.}
\end{figure}

\begin{figure}[H]
    \centering
    \begin{subfigure}{0.48\linewidth}
    \includegraphics[width=\linewidth]{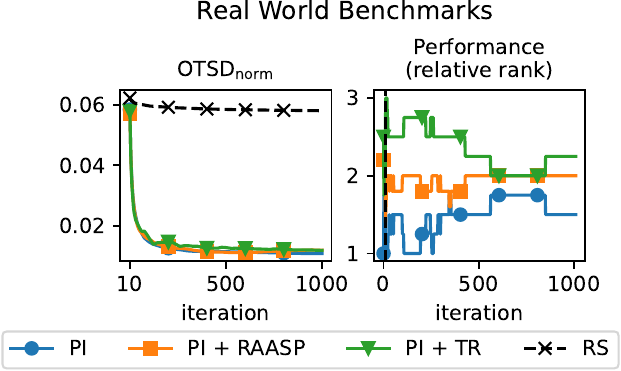}
    \caption{Without error bars.}
    \end{subfigure}
    \begin{subfigure}{0.48\linewidth}
    \includegraphics[width=\linewidth]{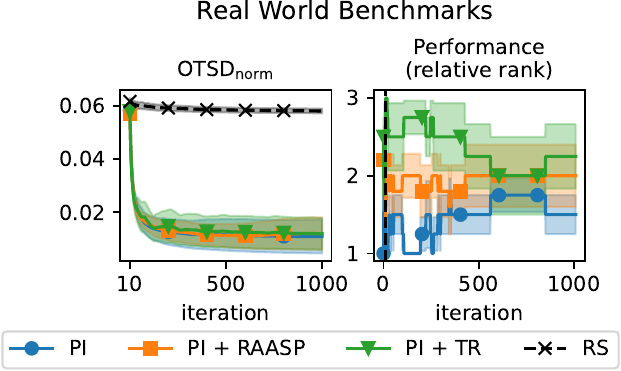}
    \caption{With error bars.}
    \end{subfigure}
    \caption{Effect of \acp{TR} and \ac{RAASP} sampling on \ac{PI} in the context of real-world benchmarks: Both methods reduce the level of exploration.}
\end{figure}

\subsubsection{Upper Confidence Bounds}

\begin{figure}[H]
    \centering
    \begin{subfigure}{0.48\linewidth}
    \includegraphics[width=\linewidth]{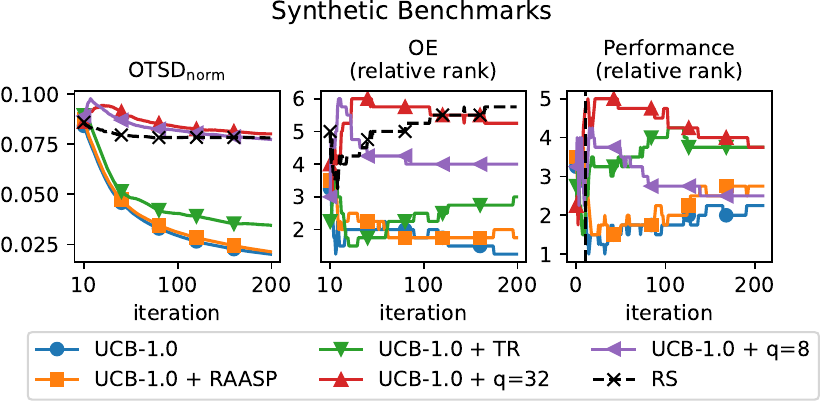}
    \caption{Without error bars.}
    \end{subfigure}
    \begin{subfigure}{0.48\linewidth}
    \includegraphics[width=\linewidth]{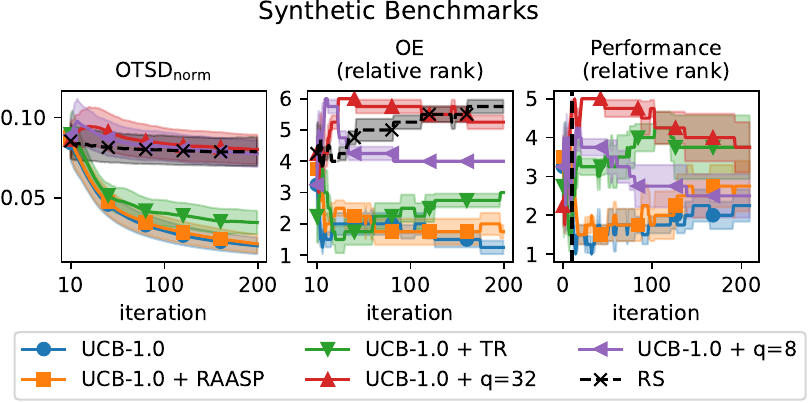}
    \caption{With error bars.}
    \end{subfigure}
    \caption{Effect of \acp{TR}, \ac{RAASP} sampling, and batching on \ac{UCB}-1 in the context of synthetic benchmarks:\ac{RAASP} sampling and \acp{TR} reduce exploration, batching increases it.}
\end{figure}

\begin{figure}[H]
    \centering
    \begin{subfigure}{0.48\linewidth}
    \includegraphics[width=\linewidth]{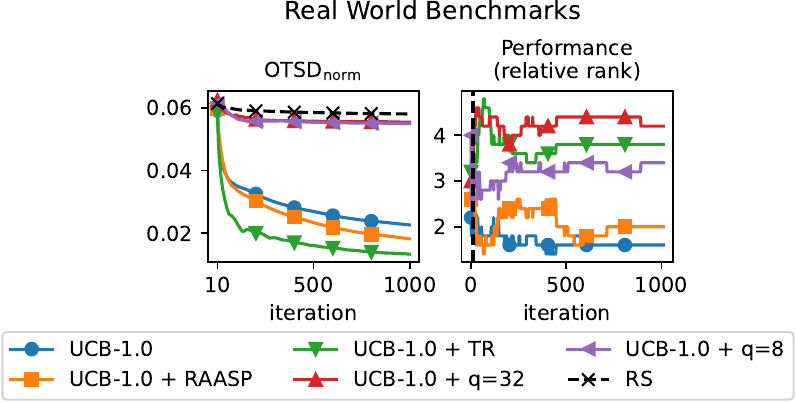}
    \caption{Without error bars.}
    \end{subfigure}
    \begin{subfigure}{0.48\linewidth}
    \includegraphics[width=\linewidth]{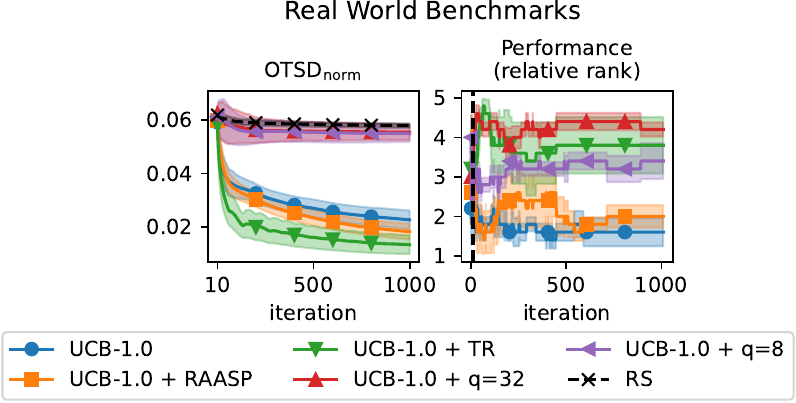}
    \caption{With error bars.}
    \end{subfigure}
    \caption{Effect of \acp{TR}, \ac{RAASP} sampling, and batching on \ac{UCB}-1 in the context of real-world benchmarks:\ac{RAASP} sampling and \acp{TR} reduce exploration, batching increases it.}
\end{figure}

\subsubsection{Thompson Sampling}

\begin{figure}[H]
    \centering
    \begin{subfigure}{0.48\linewidth}
    \includegraphics[width=\linewidth]{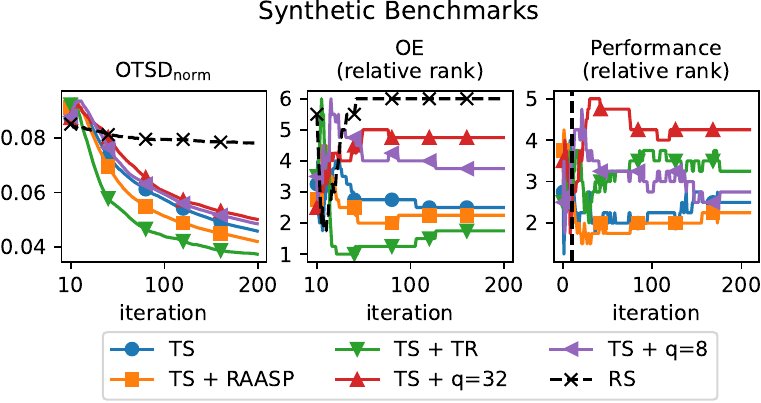}
    \caption{Without error bars.}
    \end{subfigure}
    \begin{subfigure}{0.48\linewidth}
    \includegraphics[width=\linewidth]{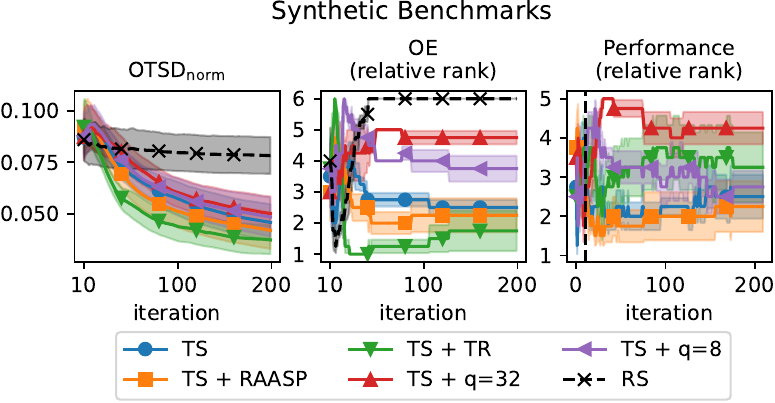}
    \caption{With error bars.}
    \end{subfigure}
    \caption{Effect of \acp{TR}, \ac{RAASP} sampling, and batching on \ac{TS} in the context of synthetic benchmarks:\ac{RAASP} sampling and \acp{TR} reduce exploration, batching increases it. \ac{RAASP} sampling also improves optimization performance.}
\end{figure}

\begin{figure}[H]
    \centering
    \begin{subfigure}{0.48\linewidth}
    \includegraphics[width=\linewidth]{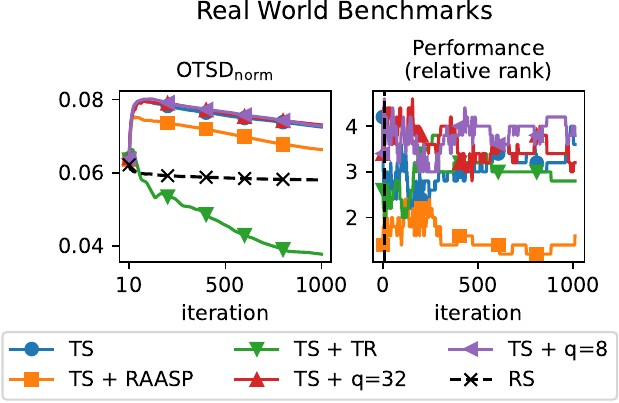}
    \caption{Without error bars.}
    \end{subfigure}
    \begin{subfigure}{0.48\linewidth}
    \includegraphics[width=\linewidth]{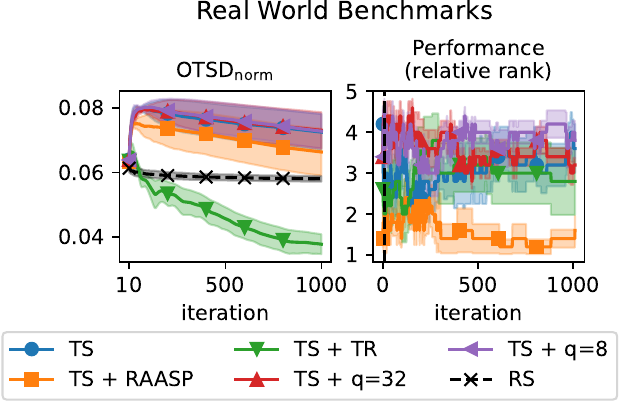}
    \caption{With error bars.}
    \end{subfigure}
    \caption{Effect of \acp{TR}, \ac{RAASP} sampling, and batching on \ac{TS} in the context of real-world benchmarks:\ac{RAASP} sampling and \acp{TR} reduce exploration, batching increases it. \ac{RAASP} sampling also improves optimization performance.}
\end{figure}

\subsubsection{Max-Value Entropy Search}

For \ac{MES}, we only study the effect of \ac{RAASP} sampling as we observed model-fitting errors for the \acp{TR}.
Furthermore, batching is not implemented for \ac{MES} in \texttt{BoTorch}.

\begin{figure}[H]
    \centering
    \begin{subfigure}{0.48\linewidth}
    \includegraphics[width=\linewidth]{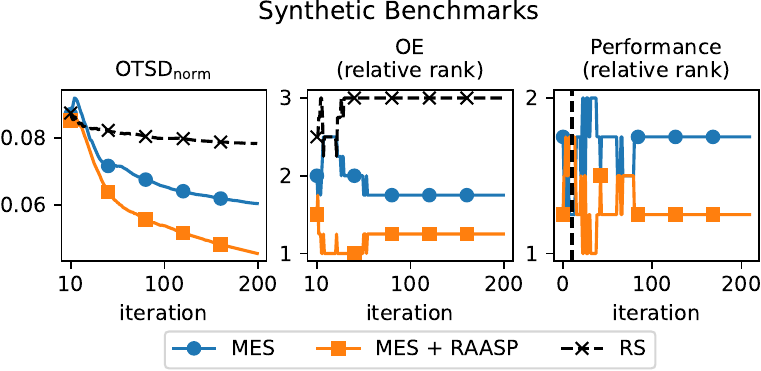}
    \caption{Without error bars.}
    \end{subfigure}
    \begin{subfigure}{0.48\linewidth}
    \includegraphics[width=\linewidth]{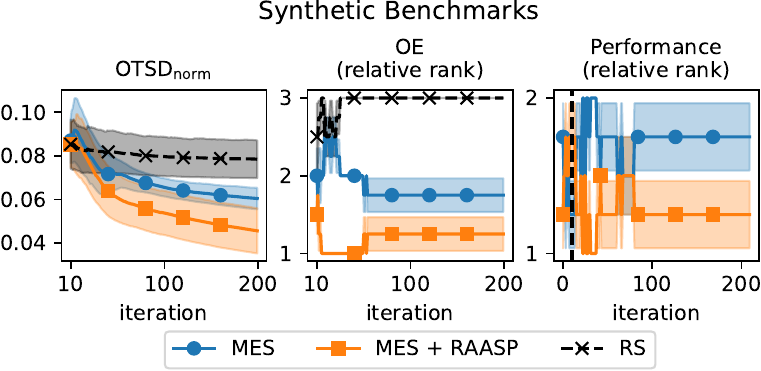}
    \caption{With error bars.}
    \end{subfigure}
    \caption{Effect of \ac{RAASP} sampling on \ac{MES} in the context of synthetic benchmarks: \ac{RAASP} sampling reduces the level of exploration.}
\end{figure}

\begin{figure}[H]
    \centering
    \begin{subfigure}{0.48\linewidth}
    \includegraphics[width=\linewidth]{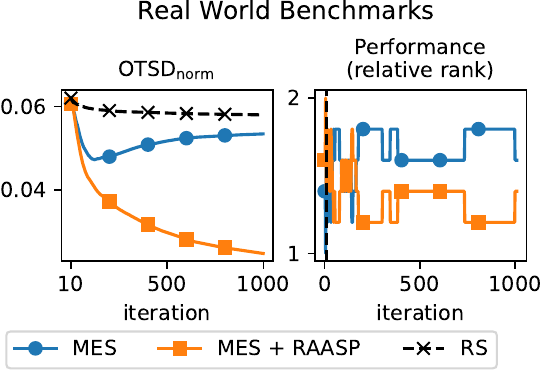}
    \caption{Without error bars.}
    \end{subfigure}
    \begin{subfigure}{0.48\linewidth}
    \includegraphics[width=\linewidth]{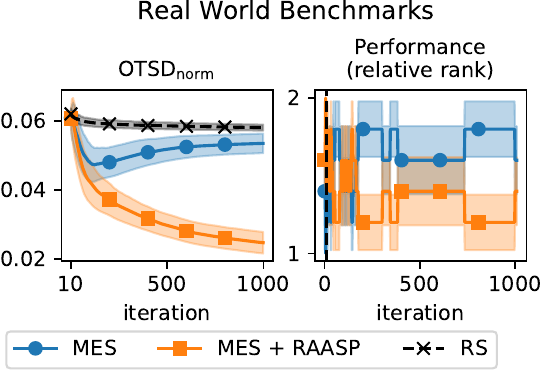}
    \caption{With error bars.}
    \end{subfigure}
    \caption{Effect of \ac{RAASP} sampling on \ac{MES} in the context of real-world benchmarks: 
    \ac{RAASP} sampling reduces the level of exploration.}
\end{figure}

\subsubsection{Knowledge Gradient}

We only ran \ac{KG} for low-dimensional synthetic benchmarks due to its high computational cost in high dimensions.

\begin{figure}[H]
    \centering
    \begin{subfigure}{0.48\linewidth}
    \includegraphics[width=\linewidth]{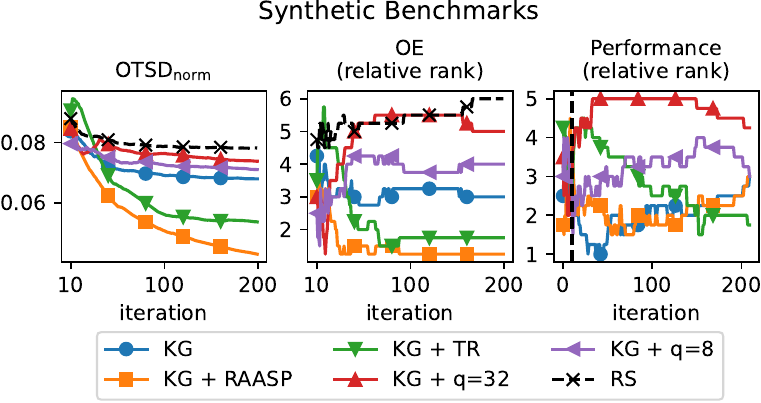}
    \caption{Without error bars.}
    \end{subfigure}
    \begin{subfigure}{0.48\linewidth}
    \includegraphics[width=\linewidth]{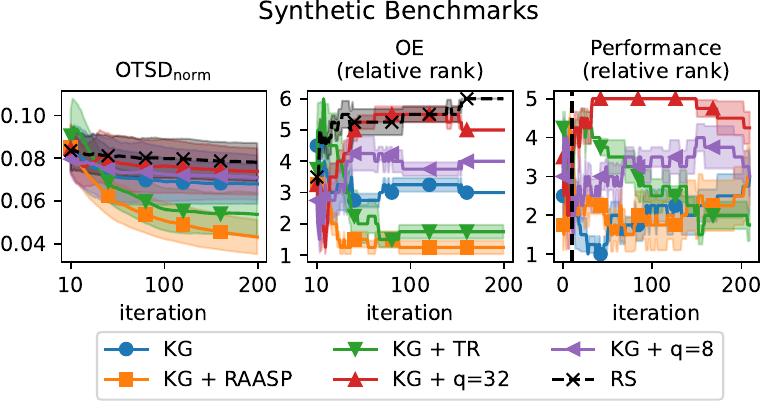}
    \caption{With error bars.}
    \end{subfigure}
    \caption{Effect of \acp{TR}, \ac{RAASP} sampling, and batching on \ac{KG} in the context of synthetic benchmarks:\ac{RAASP} sampling and \acp{TR} reduce exploration, batching increases it. }
\end{figure}

\subsection{Optimization Performance}
\label{app:raw-optimization-performance}

We present the optimization performance of various acquisition functions (\acp{AF}) and their variants for each individual benchmark. These results form the basis for the performance ranking plots in Section~\ref{sec:experiment}.
For synthetic benchmarks with known optimal values, we show the simple regret, whereas we show the best-observed function value at each iteration for the real-world benchmarks with unknown optima.

\newpage
\null
\vfill

\begin{figure}[H]
    \centering
    \includegraphics[width=\linewidth]{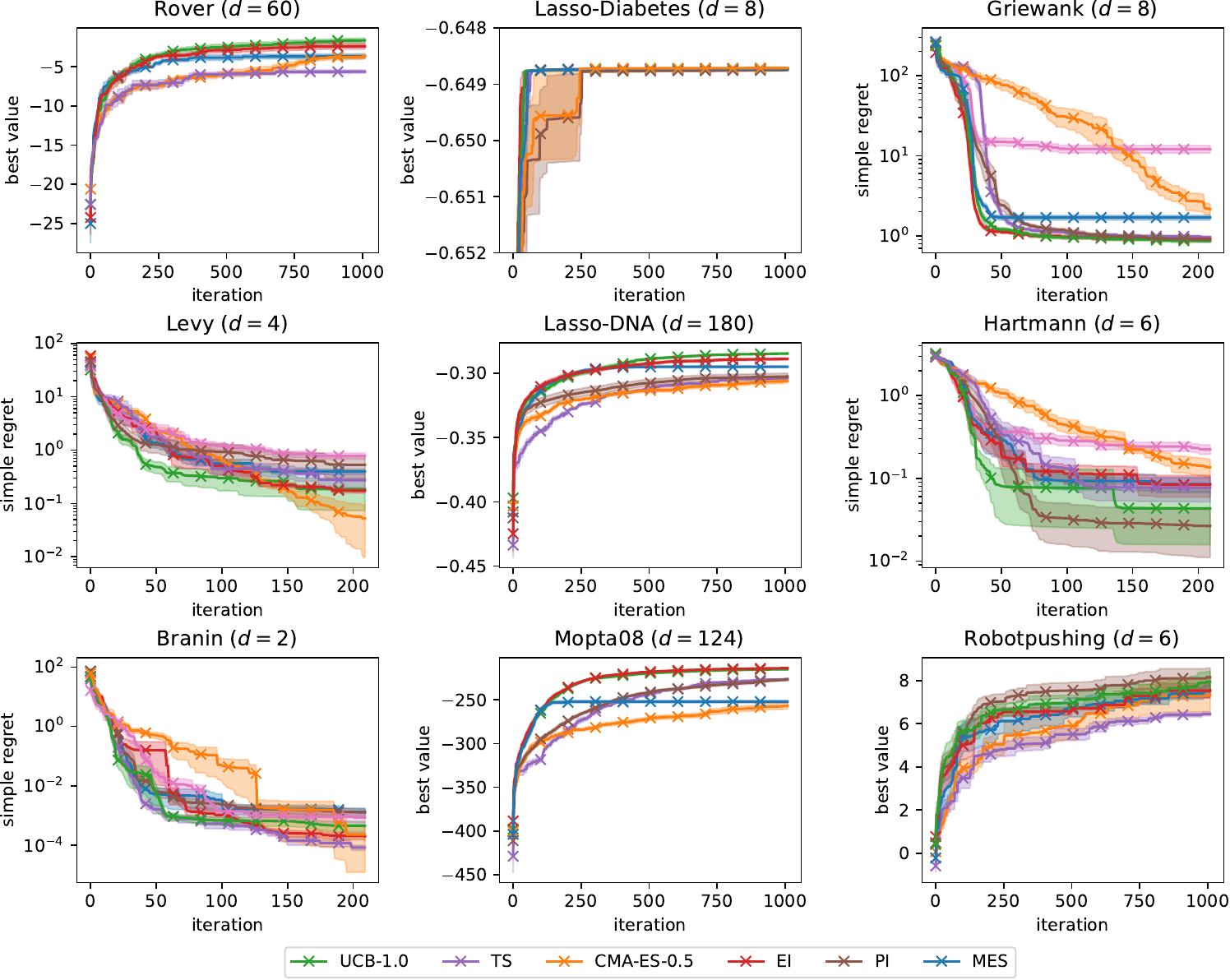}
    \caption{Optimization performance of the basic optimizer configuration on the different benchmarks.}
\end{figure}

\vfill

\newpage
\null
\vfill

\begin{figure}[H]
    \centering
    \includegraphics[width=\linewidth]{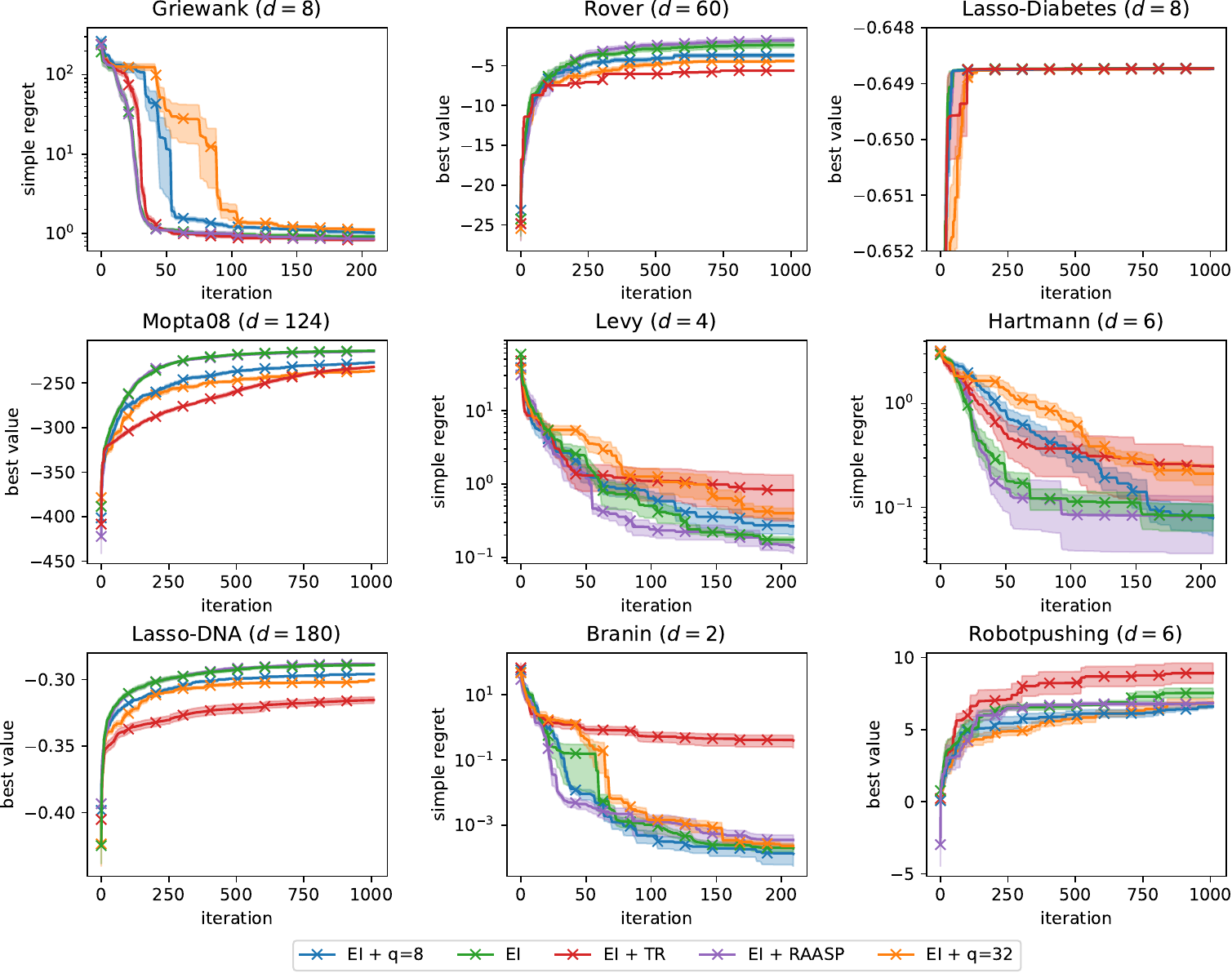}
    \caption{Optimization performance of the different EI variations on the benchmarks.}
\end{figure}

\begin{figure}[H]
    \centering
    \includegraphics[width=\linewidth]{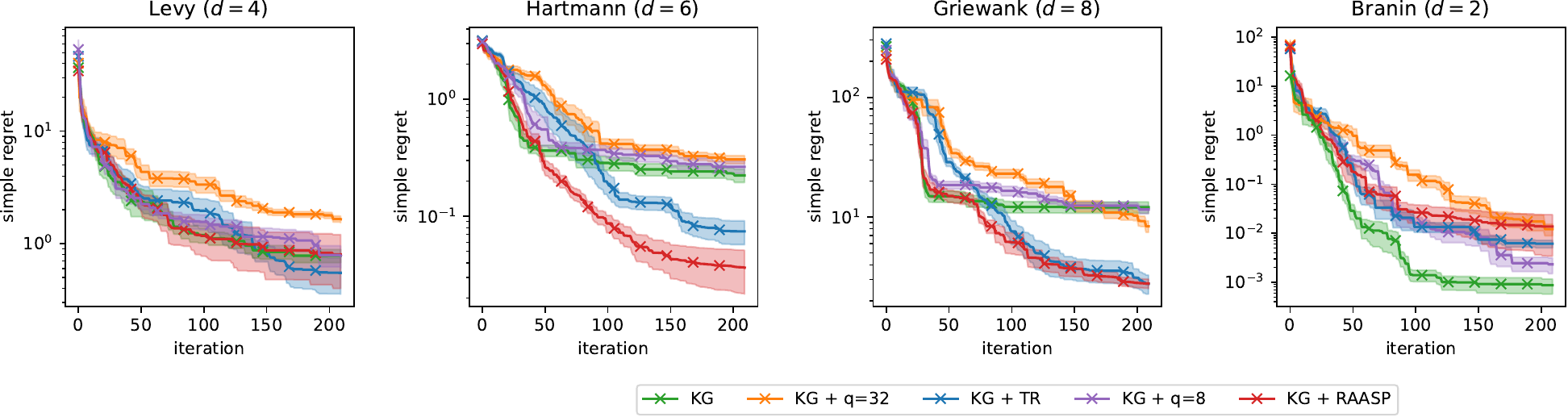}
    \caption{Optimization performance of the different KG variations on the benchmarks.}
\end{figure}

\vfill
\newpage
\null
\vfill

\begin{figure}[H]
    \centering
    \includegraphics[width=\linewidth]{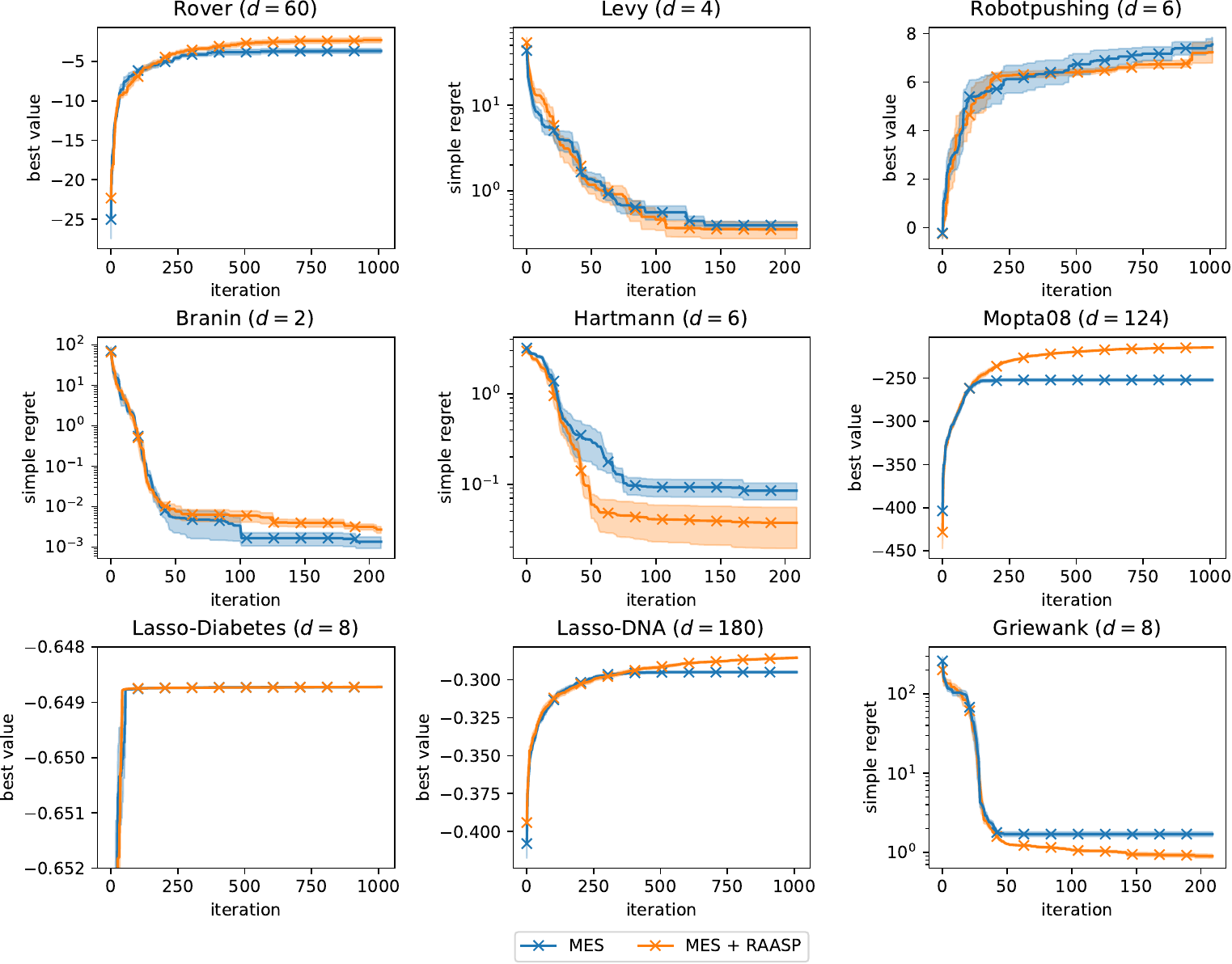}
    \caption{Optimization performance of the different MES variations on the benchmarks.}
\end{figure}

\vfill
\newpage
\null
\vfill

\begin{figure}[H]
    \centering
    \includegraphics[width=\linewidth]{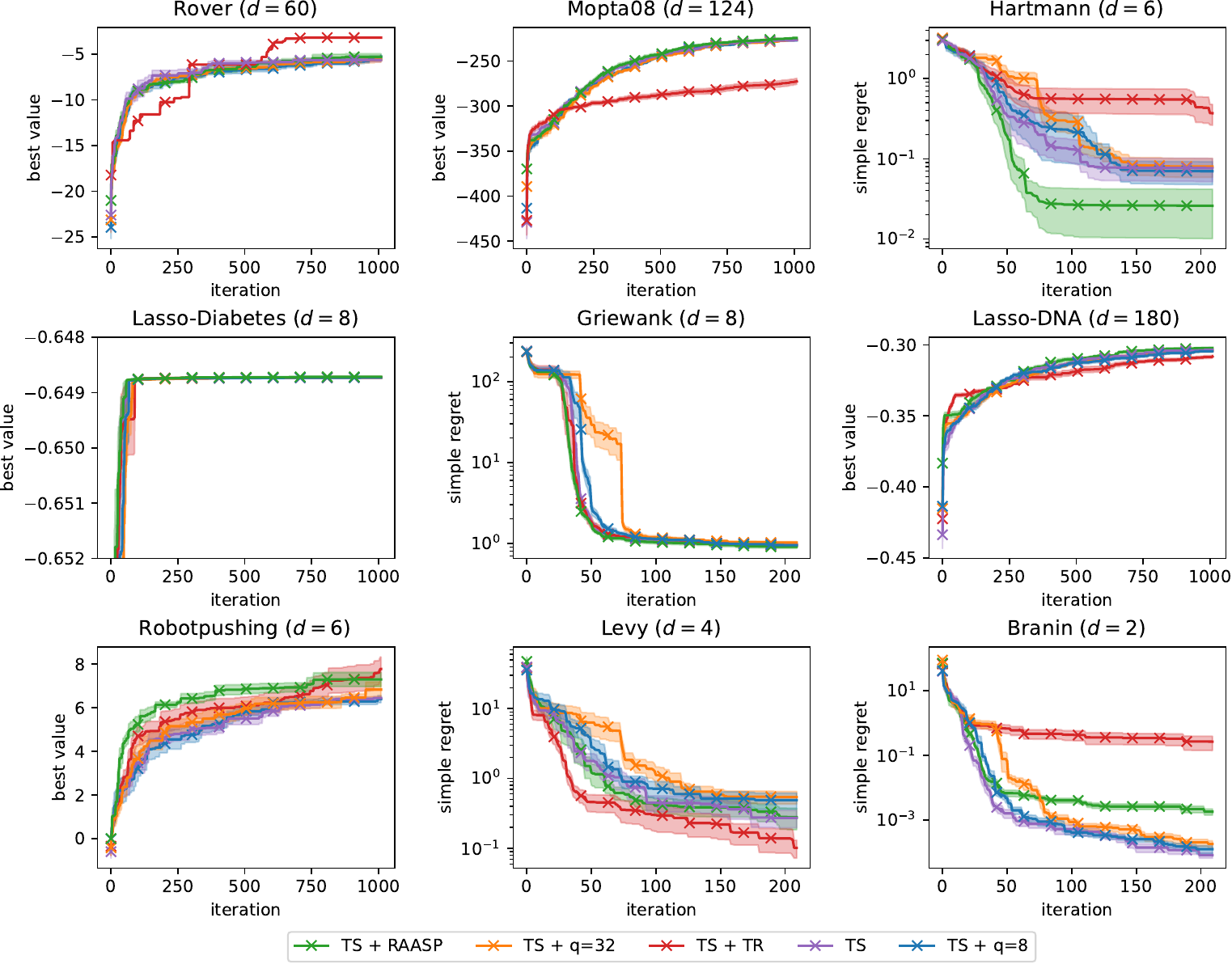}
    \caption{Optimization performance of the different TS variations on the benchmarks.}
\end{figure}

\vfill
\newpage

\begin{figure}[H]
    \centering
    \includegraphics[width=\linewidth]{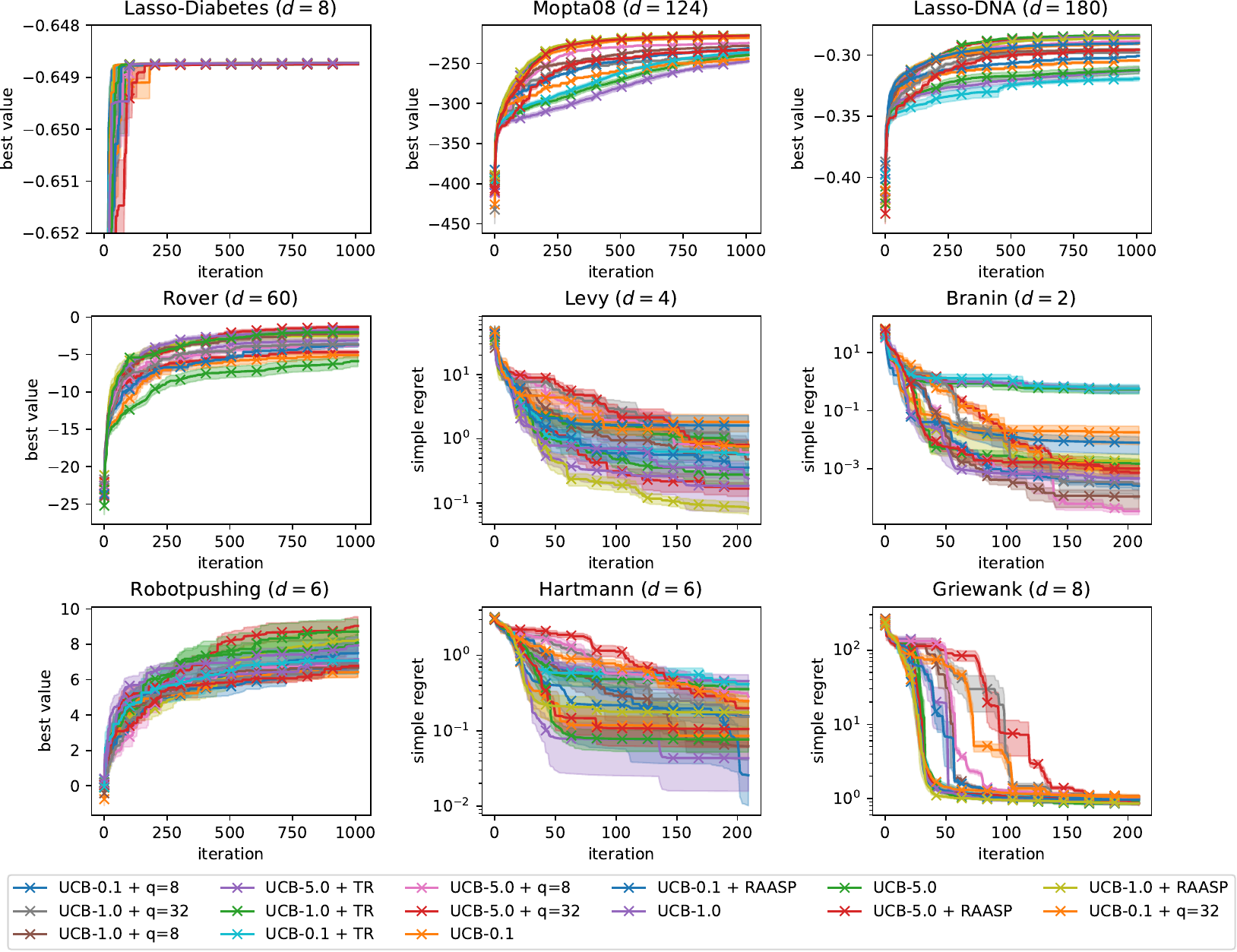}
    \caption{Optimization performance of the different UCB variations on the benchmarks.}
\end{figure}

\section{Runtime Analysis}
\label{app:runtime}

\begin{figure}[H]
    \centering
    \begin{subfigure}{.4\linewidth}
        \includegraphics[width=\linewidth]{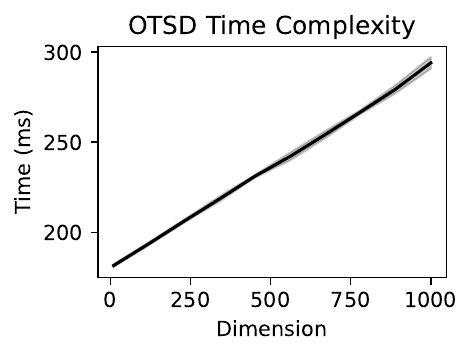}
    \end{subfigure}
    \hfill
    \begin{subfigure}{.4\linewidth}
        \includegraphics[width=\linewidth]{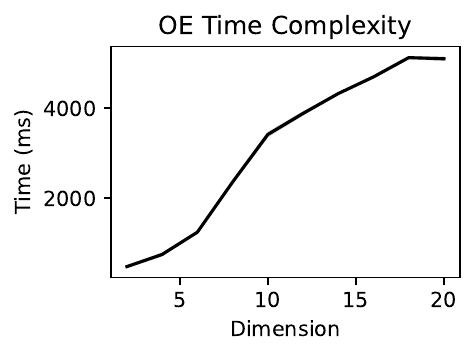}
    \end{subfigure}
    \caption{Runtimes for \ac{OTSD} and \ac{OE} in different dimensions. Empirically, \ac{OTSD} scales linearly with the number of dimensions. \ac{OE} is costlier.}
    \label{fig:runtimes}
\end{figure}

We analyze the runtimes for \ac{OTSD} and \ac{OE}.
We measure \ac{OTSD} and \ac{OE} on random sequences of length 1000 in different dimensionalities, ranging from 10 -- 1000 for \ac{OTSD} and from 2 -- 20 for \ac{OE}.
We repeat each experiment 20 times and observe the runtimes for calculating \ac{OTSD} and \ac{OE}.
Fig.~\ref{fig:runtimes} shows the runtimes for computing \ac{OTSD} and \ac{OE}. 
Computing \ac{OTSD} is fast; for 1000 observation points in 10 dimensions it takes less than 200 ms and in 1000 dimensions less than 300 ms.
\ac{OE} is considerably more costly and restricted to low-dimensional spaces.

\end{document}

%% file: math_commands.tex
\usepackage{amsmath,amsfonts,bm}

\def\1{\bm{1}}

\DeclareMathAlphabet{\mathsfit}{\encodingdefault}{\sfdefault}{m}{sl}
\SetMathAlphabet{\mathsfit}{bold}{\encodingdefault}{\sfdefault}{bx}{n}

\usepackage{amsmath,amsthm,verbatim,amssymb,amsfonts,amscd,graphicx}
\usepackage{bm, bbm, mathtools}
\usepackage{wrapfig}
\usepackage{booktabs}
\usepackage[linesnumbered,boxed,titlenumbered,ruled,nofillcomment,algo2e]{algorithm2e}
\usepackage{algorithm}
\usepackage{algorithmic}
\let\classAND\AND
\let\AND\relax

\let\AND\classAND

\usepackage[capitalize,noabbrev]{cleveref}
\usepackage{subcaption}
\usepackage{nicefrac}
\usepackage{placeins}
\usepackage{siunitx}
\usepackage{enumitem}

\theoremstyle{plain}
\newtheorem{theorem}{Theorem}[section]
\newtheorem{proposition}[theorem]{Proposition}

\theoremstyle{definition}
\newtheorem{definition}[theorem]{Definition}

\theoremstyle{remark}

\newcommand{\OTSD}{\textup{OTSD}}
\renewcommand{\OE}{\textup{OE}}